\documentclass{article}

\usepackage{arxiv}

\usepackage[utf8]{inputenc} 
\usepackage[T1]{fontenc}    
\usepackage{hyperref}       
\usepackage{url}            
\usepackage{booktabs}       
\usepackage{amsfonts}       
\usepackage{nicefrac}       
\usepackage{microtype}      
\usepackage{lipsum}
\usepackage{graphicx}
\graphicspath{ {./images/} }
\usepackage{multirow}
\usepackage[utf8]{inputenc}
\usepackage{anyfontsize}
\usepackage{csquotes}

\usepackage[
  backend=biber,
  style=authoryear,
  natbib=false
]{biblatex}
\bibliography{references}
\usepackage{colortbl}   
\usepackage{xcolor}     
\usepackage{comment}
\usepackage{booktabs}
\usepackage{tabularx}
\usepackage{array}
\usepackage{makecell}
\usepackage{listings}

\lstdefinestyle{appendixcode}{
  basicstyle=\ttfamily\scriptsize,
  breaklines=true,
  breakatwhitespace=true,
  columns=fullflexible,
  keepspaces=true,
  showstringspaces=false,
  upquote=true,
  xleftmargin=0pt,
  xrightmargin=0pt,
  aboveskip=6pt,
  belowskip=6pt
}
\usepackage{steinmetz}
\graphicspath{{./images/}}
\usepackage{amsmath}
\title{Taming the Centaur(s) with LAPITHS: a framework for a theoretically grounded interpretation of AI performances}

\author{
Matteo Da Pelo \\
  University of Cagliari, Italy\\
  \texttt{matteo.dapelo@unica.it} \\
   \And
Alessio Donvito \\
 University of Bari Aldo Moro, Italy \\
  \texttt{a.donvito22@phd.uniba.it} \\
  \And
 Claudio Frongia \\
 University of Cagliari, Italy\\
  \texttt{claudio.frongia@unica.it} \\
    \And
 Pietro Salis \\
   University of Cagliari, Italy\\
  \texttt{psalis@unica.it} \\
    \And
 Antonio Lieto \\
  University of Salerno, CIIT Lab, Italy\\
  \texttt{alieto@unisa.it} \\
}

\begin{document}
\maketitle
\begin{abstract}
We introduce a framework called LAPITHS (Language‑model Analysis through Paradigm‑grounded Interpretations of Theses about Human‑likenesS) and use it to show that several major claims advanced by models such as CENTAUR—proposed as an artificial Unified Model of Cognition—are not theoretically or empirically justified. LAPITHS provides a principled reference point for counteracting the current behaviouristic tendency in AI research to interpret the human-level performances of transformer-based language models as evidence of human-like underlying computation and, by extension, as signs of cognitive abilities.
The novelty of LAPITHS lies in making explicit the arguments grounded in two quantitative assessments: (i) the Minimal Cognitive Grid, a theoretically motivated method for estimating the cognitive plausibility of artificial systems, and (ii) a behavioural comparison showing that results similar to those reported for CENTAUR‑like models can be reproduced by other systems that do not satisfy the structural constraints typically associated with cognitive plausibility, and whose outputs do not provide independent explanatory insight into human cognition.
\end{abstract}

\keywords{Computational Cognitive Science, Cognitive Modeling, Artificial Intelligence, Cognitive Plausibility, Unified Theory of Cognition, Minimal Cognitive Grid.}



\maketitle

\section{Introduction}

\label{sec:intro}

In recent years, the rapid development of large-scale transformer-based systems has significantly reshaped the landscape of both artificial intelligence and cognitive science \parencite{sartori2023language, qu2024performance, tuckute2024language, nie2025large, wulff2026addressinglongstandingchallengescognitive, radford2019language, touvron2023llamaopenefficientfoundation}. Within this context, a growing line of research has advanced the idea that sufficiently large and properly trained language models may serve as candidates for modelling human cognition itself \parencite{binz2023using, chen2026does}. 

A paradigmatic example of this trend is Centaur, a system built through the fine-tuning of a large language model (Llama 3.1 70B) on an extensive corpus of human behavioural data, namely Psych-101, comprising over ten million decisions collected across a wide range of psychological experiments. The model is presented as a general-purpose predictor of human behaviour across multiple domains, and has been associated with the broader ambition of contributing to a unified, data-driven theory of cognition \parencite{binz2025foundation}. 

Despite these promising developments, a number of conceptual and methodological issues arise when one examines more closely the nature of the claims involved. In particular, the success of Centaur in predicting human responses across heterogeneous tasks raises a central problem: whether human-level behavioural predictive performance can legitimately be taken as evidence that the model implements cognitively plausible mechanisms. The model’s design, grounded in large-scale fine-tuning on curated behavioural data,  encourages the reproduction of human-level outputs. As such, its success may be understood as a consequence of optimisation toward behavioural agreement, rather than as evidence of shared underlying mechanisms. 

The roots of this problem can be traced to a broader tendency within contemporary AI research, namely the implicit adoption of a functionalist perspective in which matching input–output behaviour is treated as sufficient for establishing cognitive equivalence. As discussed in the literature, and explicitly addressed in this work, such a stance risks conflating behavioural adequacy with cognitive plausibility \parencite{quattrociocchi2026statistical, palminteri2026, lerchner2026abstraction}. 

In order to address this problem, the present work introduces a theoretically grounded framework—LAPITHS (Language models Analysis through Paradigm-grounded Interpretations of Theses about Human-likeness)—designed to disentangle performance from cognitive interpretation.\footnote{Lapiths was also the name of an ancient population fighting the Centaurs in the epic narration of Centauromachy, sculpted in the Parthenon.} The framework combines a conceptual analysis of the conditions under which cognitive ascriptions are warranted with an operational evaluation grounded in the Minimal Cognitive Grid (MCG) \parencite{lieto2021cognitive}. 
The MCG provides a structured methodology for assessing artificial systems along three key dimensions: functional/structural ratio, generality, and performance match. Crucially, it treats behavioural success as just one component of a broader evidential framework, allowing for a more detailed assessment of cognitive plausibility. 

The empirical component of this work applies the LAPITHS framework to a series of behavioural and representational analyses centred on the two-step task, a canonical paradigm in reinforcement learning research \parencite{daw2011model} and a crucial task for assessing claims made about the model (since its performances have led to conclusions about both neural and behavioural alignment by the authors, see Section \ref{sec:fmri_test} for details). The results indicate that other state-of-the-art language models, not specifically trained on the task to handle but only informed via RAG about the structure of the task, are capable of achieving levels of behavioural fit comparable to those reported for Centaur, with differences that are in some cases statistically not significant. Moreover, representational analyses based on fMRI data suggest that high levels of alignment with human neural activity can be reproduced by models that have not undergone task-specific fine-tuning. 

Taken together, these results support a more cautious interpretation of recent developments. While models such as Centaur demonstrate some strong predictive capabilities, their performance appears to reflect functional adequacy at the level of input–output behaviour rather than the implementation of cognitively plausible mechanisms \parencite{quattrociocchi2025epistemological}. The LAPITHS framework, and in particular the MCG analysis, makes explicit the extent to which such systems diverge from the structural constraints identified in cognitive theory, thereby providing a principled basis for evaluating their explanatory status.

The structure of the paper is organised as follows: Section \ref{sec:centaur_intro} introduces Centaur in detail, outlining its architecture, training methodology, and the composition of the Psych-101 dataset. Section \ref{sec:funct_VS_struct} develops the theoretical foundations of the critique, focusing on the distinction between functional and structural models, the role of functionalism in AI, and the formulation of the ascription fallacy. Section \ref{sec:framework} presents the LAPITHS framework, including the formalisation of the MCG and its application to the evaluation of artificial systems. Section \ref{sec:results} reports the empirical analyses, including behavioural replication experiments on the two-step task, statistical comparisons with baseline models, and an fMRI-based representational study. Finally, Section \ref{sec:discussions} discusses the broader implications of the findings for the evaluation of AI systems and for the project of cognitive modelling, highlighting directions for future research.

\section{What Is Centaur}\label{sec:centaur_intro}
Centaur is a foundation Artificial Intelligence model based on Llama 3.1 70B \parencite{bommasani2021opportunities, touvron2023llamaopenefficientfoundation, grattafiori2024llama}. 
In the author's view, a crucial step toward a unified theory of cognition is the development of a computational model capable of predicting and simulating human behaviour across all domains \parencite{wu2018generalization}. Based on the reported data, the authors consider Centaur as a candidate for a unified theory of cognition \parencite{newell1973twentyquestions, newell1994unified, anderson2003newell, laird2017standard} proposing the idea that such behavioural replication is an indicator of some underlying cognitive alignment between the model and the mechanisms of human cognition. Centaur, in fact, appears to effectively predict the decisions made by some human participants in the Psych 101 test dataset.

The dataset was constructed by transcribing data from 160 psychological experiments into natural language. Each prompt was designed to include the entire sequence of a complete session for a single participant in each trial. The transcription of each experiment was manually performed by the authors. The maximum length of each prompt was set at approximately 32,768 tokens. The dataset includes six categories of experiments, covering: (i) decision-making, (ii) memory, (iii) supervised learning, (iv) Markov decision processes, (v) multi-armed bandits, (vi) miscellaneous. 
Following the creation of this dataset, Centaur was trained through the fine-tuning of Llama specifically on Psych 101. Fine-tuning is a technique for adapting a model’s initial dataset to a specific knowledge domain. This process also involves backpropagation and weight optimization, but it operates on a significantly smaller dataset compared to pre-training. Fine-tuning enables the model to achieve higher performance on task-specific applications. The fine-tuning technique employed for Centaur is QLoRA, or Quantised Low-Rank Adaptation. QLoRA preserves the base model’s parameters, thus trying to freeze as much as possible the representations obtained during the training phase, while introducing low-rank adapters containing a small number of additional trainable parameters. This approach enhances the well-known LoRA method, achieving comparable performance while significantly reducing energy and memory consumption. This efficiency improvement appears to be QLoRA’s primary contribution to the state of the art  \parencite{hu2021loralowrankadaptationlarge, dettmers2023qlora}.

As a fine-tuned system, Centaur performs remarkably well on the “cognitive decathlon”, a benchmark introduced by the authors to evaluate the capacity of computational models to predict human behaviour. One of the most striking aspects of the model is that its internal representations appear to become increasingly aligned with human behavioural patterns during training. In other words, the model’s internal representations seem to synchronize with patterns extracted from the behavioural data.


Despite these noteworthy achievements, several theoretical concerns arise regarding Centaur. In particular, some aspects of the model’s design raise questions about its suitability as a robust cognitive predictor. Moreover, certain experimental findings suggest that the reported results may be replicable with other approaches without any explicit commitment to explanatory adequacy. Our analysis suggests that evaluating the performance of a fine-tuned large language model in experimental contexts requires a careful examination of the training dataset itself. The dataset may contain implicit properties or latent regularities that contribute to the model’s apparent success. Consequently, understanding how the training data are constructed and organized becomes crucial for explaining how the model achieves its performance and for assessing the extent to which its results genuinely reflect cognitive modelling rather than dataset-specific learning effects \parencite{gluck2018interactive}. In the following paragraph, a fundamental methodological distinction in AI design is introduced, which underpins one of our central objections to the results and overall design of Centaur.

\section{Functional vs. Structural Artificial Models}\label{sec:funct_VS_struct}

An important methodological distinction in AI and computational cognitive modelling is that between functionally and structurally designed models. This distinction is central to the debate in the explanatory role played by artificial models with respect to the natural cognitive systems that they take as sources of inspiration. 

Functionalism was introduced into the philosophy of mind by Hilary Putnam, beginning with his influential paper “Minds and Machines” \parencite{putnam1960minds}, as an alternative to type‑identity theory based on the idea that mental states are individuated by their functional roles rather than by their physical realisations. In particular, two tokens are assumed to belong to the same mental state if they stand in the same functional relations to other mental states and to the system’s inputs and outputs. Functionalism led, in AI, to the definition of a design approach based on the notion of “functional equivalence” between some cognitive faculties (to be modelled) and the corresponding mechanisms implemented in AI programs \parencite{lieto2021cognitive}. Indeed, its more radical formulation postulated the sufficiency, from an epistemological perspective, of a weak equivalence between cognitive processes and AI procedures, positing that, from an explanatory point of view, the relation between “natural mind” and “artificial software” could have been based purely on a macroscopic input-output equivalence. In the last decades, this position has been widely criticised. In particular, models and systems designed according to the “functionalist” perspective have been deemed inadequate for advancing the science of cognitive AI.\footnote{\textcolor{black}{Functionalism does not necessarily entail computationalism, and the possibility of non‑computable mental functions remains a live philosophical option. The distinction between functional and structural models adopted here is restricted to computationally realizable models and does not presuppose that every mental function is computable. If some cognitive functions were non-computable, they would fall outside the framework insofar as no computational model could reproduce them exactly, although computable components or approximations might still be modelled. This broader possibility does not affect the present analysis, which is explicitly framed within the domain of computationally implementable models.}}

As in the case of an airplane—which flies through mechanisms entirely different from those of a bird \parencite{russell1995artificial}—the mechanisms and design choices underlying such artefacts prevent them from playing a genuine explanatory role with respect to their analogous natural systems. This applies, for example, to all the main contemporary AI systems, including so-called “cognitive computing” technologies such as IBM Watson \parencite{ferrucci2010building}, AlphaGo \parencite{silver2016mastering}, and large language models like GPT-5 or DeepSeek R1 \parencite{guo2025deepseekr1}. Despite the propaganda, such systems cannot be qualified as “cognitive” under a structurally grounded notion of cognition, insofar as they lack explanatory relevance with respect to: (i) how humans organise, retrieve, and reason over stored information when answering questions, or (ii) how humans plan and make decisions, etc. In other words, most current AI systems are “functionalists”: they function as if they were natural cognitive systems in terms of output, but the internal mechanisms generating such outputs differ from those employed by humans. Therefore, artificial imitation of cognitive capacities does not imply implementation of the same underlying principles. As mentioned, this is an important aspect to point out, as there is a persistent tendency to attribute cognitive explanations to purely functional systems, a tendency reinforced by the loose and often misleading use of terms such as “cognitive computing”, which is frequently employed as a broad label for systems capable of interacting with humans. 

In contrast to functionalism, a structural approach to AI design calls for a more strongly constrained correspondence between artificial systems (their internal architectures and processes) and their natural counterparts. According to this view, structurally constrained artificial models and systems can be useful both to advance the science of AI in terms of technological achievements and to play the role of “computational experiments”, able to provide insights and results useful in refining or rethinking theoretical aspects concerning the target natural system used as a source of inspiration \parencite{milkowski2013explaining, cordeschi2002discovery}. 
 
A problem arising in this view is that it is not possible to build a completely structural and constrained artificial model, since an exact replica of a natural system is unattainable. The pursuit of increasingly structural models leads to an asymptotic regress to the microscopic physical world, culminating in the well-known Wiener paradox: “The best material model of a cat is another, or preferably the same, cat” \parencite{rosenblueth1945role}. This paradox highlights the necessity of constructing proxy models rather than replicas. Similarly, Zenon Pylyshyn \parencite{Pylyshyn1979-PYLCAT} noted that: “if we do not formulate any restriction about the model, we obtain the functionalism of the Turing machine, and if we apply all the possible restrictions, we reproduce a whole human being”. The key issue, therefore, is identifying the appropriate level of description and enforcing constraints at that level, in order to achieve human-like computation \parencite{marr2010vision}. In this view, the functionalism/structuralism dichotomy is better understood as a continuum. Between the explanatory inadequacy of pure functional artificial models (which may nonetheless achieve impressive performance in specific tasks) and the impracticality of pure structural models, there exists a spectrum of plausible artificial proxy models with varying degrees of explanatory power. Having clarified these preliminary points, we may now turn to the central focus of the article: the problems that emerge from a close reading of the scientific presentation of Centaur. A critical examination reveals a number of issues, some of which are not unique to this work but reflect broader trends in the literature, extending beyond the specific contribution of Binz and colleagues.

\subsection{Ascription Fallacy}

A central problem is the so-called “ascription fallacy”: i.e., conflating human-level behavioural performance with human-like computational mechanisms, and then explaining AI outputs using the same biological or cognitive theories we use to explain analogous human outputs \parencite{quattrociocchi2025epistemological}. Centaur is a salient case where this conflation can occur, for three connected reasons. First, Centaur’s training objective structurally encourages behavioural isomorphism, not mechanistic homology \parencite{orr2025wronglimitspredictionexplanation}. Centaur is trained to reproduce or predict human trial-by-trial responses in psychological tasks, via standard language-model fine-tuning (parameter-efficient adaptation on a pre-trained LLM). As a result, strong behavioural fit is unsurprising, since the optimisation target is explicitly defined in terms of agreement with observed human responses. However, fit-to-behaviour alone is not evidence that the system implements the same competence in a biological or cognitive sense.\footnote{\textcolor{black}{This criticism of the inference from functional equivalence (behavioural output) to structural equivalence (mechanistic identity) does not imply that we think structural similarity is sufficient to attribute cognition to a system. We are going to defend a more structured approach, according to which both functional and structural aspects are necessary for a realistic cognitive plausibility of an artificial system. Such a structured approach is the Minimal Cognitive Grid. See Section \ref{sec:framework}.}} It is evidence of functional adequacy at the interface, namely that a mapping from task description and history to response can be learned. In the “ascription fallacy” framing, the mistake occurs when this kind of functional match is treated as if it licensed a structural (mechanistic or biological) attribution.

\begin{figure}
\centerline{\includegraphics[width=0.7\textwidth]{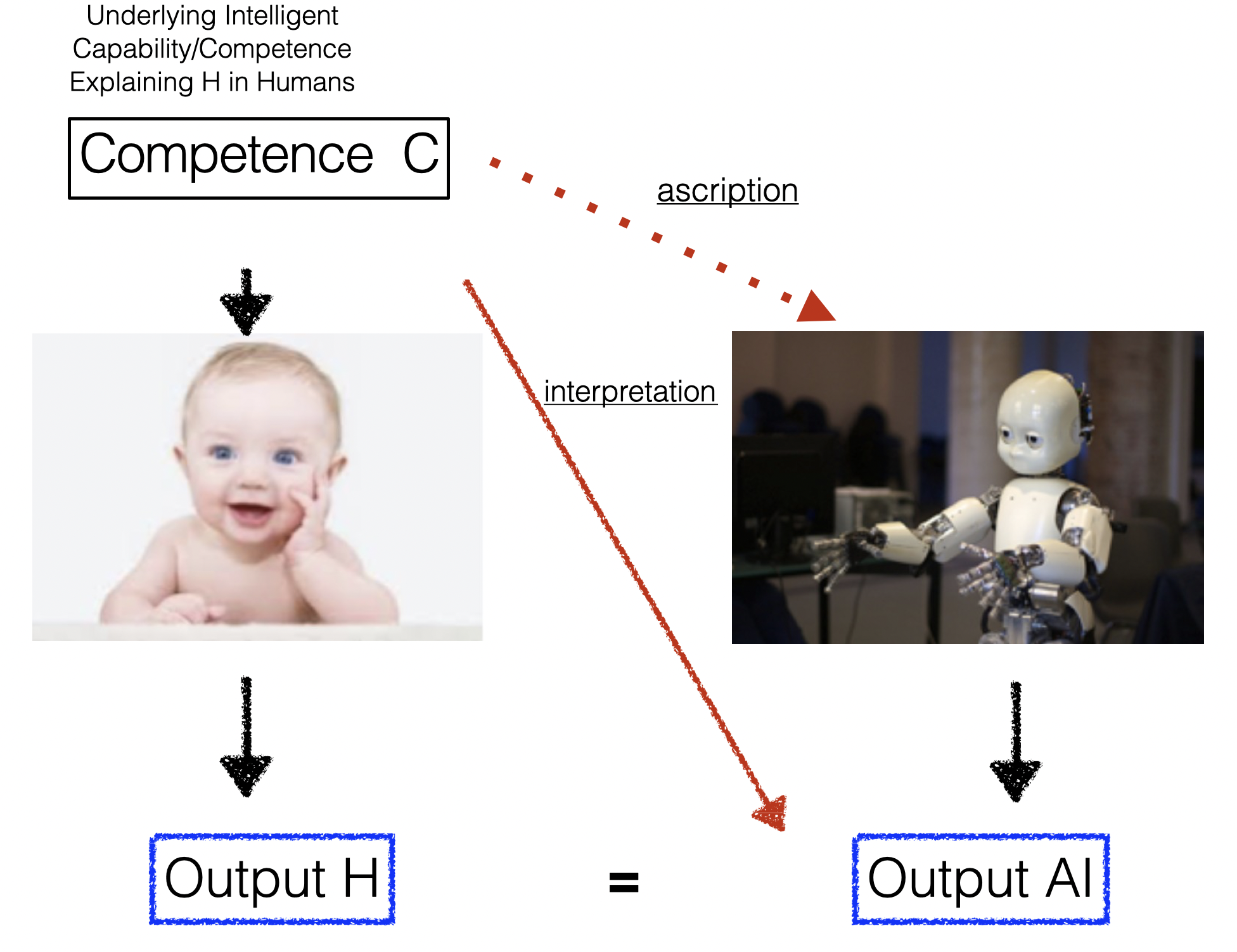}}
\caption[width=0.4\textwidth]{A pictorial representation of the ascription fallacy problem, where the fact that the Humans and AI produce the same output, despite using structurally different mechanisms, leads to the unjustified ascription, to the AI system, of the theory of competence explaining the behaviour in the natural system.}
\label{fig:fallacy}
\end{figure}

In particular, this happens when the competence of a given cognitive system for a particular task is assigned also to an artificial system for the mere equivalence of the obtained results, despite the structural differences of the mechanisms used by the natural and artificial systems to perform that task. Such ascription leads to the interpretation of the results of the artificial system in terms of the competence explaining the behaviour in the natural system (see Figure~\ref{fig:fallacy}).

Second, the inference from accurate behavioural prediction to the possession of cognition is underdetermined. Centaur’s headline contribution is broad behavioural prediction across many experiments and generalisation to certain distribution shifts. However, behavioural equivalence classes are large: many distinct internal organisations can implement the same input–output mapping, or match the same benchmark statistics, without sharing the causal or mechanistic properties that make a cognitive capacity in humans. This is exactly the “behaviourist trap” that the ascription fallacy critique targets \parencite{li2026ghost}. Once evaluation privileges outputs alone, systems that exploit statistical regularities, retrieval, or shallow heuristics can be misread as instantiating the same competence humans deploy \parencite{quattrociocchi2025epistemological}. Centaur’s success, therefore, establishes, at most, that a language-model-based architecture can serve as a high-coverage behavioural emulator for tasks encoded in natural language, not that it realises the same competence type as biological cognition. Third, “neural alignment” results are vulnerable to a reverse-inference leap. The Centaur paper reports that internal representations become “more aligned with human neural activity” after fine-tuning \parencite[p.~5]{binz2025foundation}. This type of result is often rhetorically taken to support a mechanistic reading, suggesting that the model is becoming brain-like. Methodologically, however, representational similarity is correlational and level-ambiguous. It can arise because the model and humans are exposed to similar task structures, stimulus descriptions, and response demands, without implying shared computational primitives, learning dynamics, or causal decompositions into human cognitive mechanisms. The ascription fallacy is triggered if one moves from “some representational alignment under a given analysis” to “therefore, the system instantiates the same cognitive process”, especially when this move is used to explain human cognition in terms of the model’s learned internal features.

\subsection{Centaur is functionalist by default}
The proponents of Centaur, viewing it as a promising candidate for a foundational model of cognition, underscore the importance of its performance aligning with a broad spectrum of human results (i.e., Psych-101). This raises an important consideration. The comparison between Centaur’s outcomes and human results in the dataset predominantly focuses on matching performance outputs. Consequently, a functionalist perspective on these models and results is implicitly assumed, emphasising behavioural outputs. However, the function performed by a system $S$ is distinct from the ways in which that function is implemented. Two systems, $S$ and $S_1$, may realise the same cognitive function $F$ while relying on entirely different architectures and mechanisms. Consequently, a genuinely cognitive system and a non-cognitive one may be evaluated as equivalent if assessment is based solely on behavioural outputs, regardless of whether the process involves actual cognition. For a more robust evaluation, artificial systems should be assessed along two dimensions: (1) functional design, concerning the performance of $S$ in a task $T$, and (2) structural design, addressing the architecture and mechanisms that realise that performance $T$ \parencite{lieto2021cognitive, lieto2022analyzing}. This alternative perspective has broader implications, such as the adoption of a minimal cognitive grid (MCG), comprising a basic set of functional and structural criteria to assess the cognitive profile of artificial systems \parencite{webb2001can, lieto2021cognitive, lieto2022analyzing}. 
By contrast, Centaur’s proponents appear to assume functionalism as the default stance without explicitly justifying it. Is this assumption warranted? Can we legitimately infer the human-likeness of systems such as Centaur on the basis of behavioural similarity alone? Perhaps this stance hinges on the shift from traditional symbolic artificial systems to neural networks. The argument suggests that this transition addresses the issue of structuralism at the outset, given the alleged resemblance between neural networks and the human brain. However, such resemblance is superficial.  \textcolor{black}{Contemporary neural networks differ in fundamental ways from biological neural systems such as human neural architectures.} Mechanisms such as backpropagation, attention, and memory operate according to principles that diverge significantly from those observed in the brain \parencite{crick1989recent, lillicrap2020backpropagation}. Thus, merely assuming that structuralism has been addressed through the shift from symbolic AI to neural networks is insufficient  \parencite{orr2025wronglimitspredictionexplanation, quattrociocchi2025epistemological}. We are once again faced with the necessity of critically examining the functionalist assumptions advocated by Centaur’s proponents. If its empirical results are to support ambitious claims about cognition, those assumptions must be explicitly defended. This simplistic functionalist assumption closely resembles a radical version of functionalism akin to behaviourism, where only behavioural outputs matter. As we will show in the following sections, this assumption is problematic.

\section{The Framework}\label{sec:framework}

This section specifies a two-part methodological framework designed to test—and, where warranted, reject—the inference from linguistic output quality to possession of a cognitive competence in the human/biological sense. The motivating concern is underdetermination: many heterogeneous systems can generate the same (or indistinguishably similar) input–output behaviour yet differ radically in internal organisation, causal structure, and explanatory relevance to human cognition. The framework treats output-based success as just \textit{one component} of an evidential package, and it makes explicit what kind of additional evidence is required for a justified competence attribution. The framework has two tightly coupled components: (1) a theoretical component: a principled argument that output evidence is not identifiable with cognitive competence, plus an operational analysis of what counts as stronger evidence using a formalisation of the Minimal Cognitive Grid (MCG); (2) an experimental component (behavioural equivalence demonstrations): empirical constructions showing that the same outputs (or the same benchmark-level performance) can be produced by non-cognitive systems, and that this fact alone does not warrant cognitive ascription to those systems, revealing a flaw in output-only inference.

\subsection{Minimal Cognitive Grid}

To address the long-standing challenge of providing computational explanations of cognition \parencite{milkowski2013explaining}, the Minimal Cognitive Grid (MCG) \parencite{lieto2021cognitive} was introduced as a framework for evaluating the epistemological and explanatory status of artificial systems. Its primary aim is to assess the extent to which such systems align with relevant cognitive theories of the modelled phenomenon. Building on Webb’s seminal proposal for characterising bio-inspired models \parencite{webb2001can}, and in line with the principles of Psychometric Artificial Intelligence \parencite{bringsjord2003artificial, bringsjord2011psychometric}, Lieto developed a model-agnostic evaluation approach grounded in the following dimensions: (1) Functional/Structural Ratio: this dimension concerns the analysis of the balance between functional and structural design components in the model. In complex artificial systems, it is possible to represent certain components in a purely functional manner, especially when their internal mechanisms are not central to the system’s intended explanatory or pragmatic goal. Conversely, other components may require a structurally faithful modelling approach;
(2) Generality: it captures the extent to which a system is capable of addressing diverse cognitive domains. A system that covers a broader range of tasks is considered more cognitively plausible, as it better reflects the versatility typical of human cognition;
(3) Performance Match: this dimension requires a direct comparison between artificial and biological/cognitive systems, focusing on their performance across specific tasks through the use of standardised and validated benchmarks to assess the degree of alignment between artificial behaviour and human cognition. Two additional criteria are proposed to refine the evaluation: (a) error pattern analysis: the system should exhibit error types and distributions comparable to those observed in human performance; (b) execution time: the time taken by the artificial system to complete tasks should approximate human timing under analogous conditions.
Taken together, the three dimensions of the Minimal Cognitive Grid provide a non-subjective method for assessing the degree of cognitive plausibility of computational models. In this contribution, we adopt a model-agnostic formalisation of the MCG dimensions \parencite[as proposed in][]{donvitolieto2026}, clarifying how it can be operationalised independently of specific domains (e.g., analogy, language, vision) or reference models. The framework is intended as a reusable methodological tool for quantitative model analysis.

\subsubsection{Functional/Structural Ratio}

In this section, we present the metric to measure the ratio between functionality and structurality in the target computational models. The metric, which we call $FSR_\mathcal{M}$, allows us to evaluate how a given computational model $\mathcal{M}$ adheres to one or more selected cognitive theories of the modelled phenomenon. First of all, given cognitive theories $\{T_1, \dots, T_n\}$, a set of $k$ structural constraints $\{c_1, \dots, c_k\}$ must be derived. These represent explicit theoretical features derived from cognitive science that specify how certain cognitive functions are internally realised, and may include processing assumptions, mechanism-specific features (e.g., the use of working memory buffers), learning mechanisms, modularity, etc. Once $\{c_1, \dots, c_k\}$ have been identified and appropriate criteria for observing their implementation within computational models have been defined, each model $\mathcal{M}$ can be evaluated against the full set of constraints \( \{c_1, \dots, c_k\} \). For each constraint \( c_i \), the evaluation involves two components: (1) Partial structural score \( s_i \in \{0, 1\} \), indicating whether the model implements the structural mechanism posited by the constraint:  
\[
  s_i =
  \begin{cases}
    1 & \text{if $\mathcal{M}$ implements the structural mechanism} \\
    0 & \text{if $\mathcal{M}$ does not implement the mechanism structurally};
  \end{cases} 
\]
(2) Weight \( w_i \in [0, 1] \), reflecting the relative importance of constraint \( c_i \) within the theory, with the normalisation condition \( \sum_i w_i = 1 \).

\noindent
Since we consider functionality as the complement of structurality, we define the partial functional score simply as:  
\begin{equation}
f_i = 1 - s_i
\end{equation}
This means that if a constraint is structurally implemented (\(s_i = 1\)), it cannot be considered functionally satisfied at the same time, and vice versa. \textcolor{black}{Here, complementarity $(f_i=1-s_i)$ is intended in an operational sense at the level of individual constraints, and not as the general claim that functional adequacy and structural fidelity are logically incompatible. A functional score, in our convention, denotes a \textit{functional-only} implementation, in which the relevant function is reproduced without implementing the mechanism specified by the reference cognitive theory.}

\noindent We can then compute the overall Structural and Functional scores (${S_{M}}, {F_{M}}$) as:
\begin{equation}
  S_M = \sum_{i=1}^{k} w_i s_i, 
  \quad
  F_M = \sum_{i=1}^{k} w_i f_i = 1 - S_M
\end{equation}
And define the $\boldsymbol{FSR_\mathcal{M}}$ as:
\begin{equation}
  {FSR_\mathcal{M}} = \frac{F_M}{S_M + \varepsilon}
  = \frac{1 - S_M}{S_M + \varepsilon}
  \hspace{1cm}
\end{equation}
where $\epsilon$ is a small positive constant ($0.01$) added to prevent division by zero. Values of ${FSR_\mathcal{M}}<1$ indicate a high degree of structural plausibility, whereas values $>1$, and progressively more so as they increase, indicate that the system deviates from the internal constraints posited by $T$.

\subsubsection{Generality}

Generality measures the breadth of cognitive functions a system can replicate across various domains (e.g., language, vision, reasoning, etc). 
In order to ground our evaluation of generality in an empirically validated cognitive framework, we can select our task domains by drawing from the Cattell–Horn–Carroll (CHC) theory of cognitive abilities \parencite{carroll1993human, mcgrew2005cattell}. 
The CHC psychometric model is the most thoroughly researched, empirically validated, and comprehensive hierarchical taxonomy of human cognitive abilities, organised into three strata: 1. domain-independent capacities; 2. a middle layer of broad abilities; 3. numerous narrow abilities. Among the most relevant areas, CHC includes: Fluid Intelligence (Gf), Quantitative Knowledge (Gq), Reading-Writing (Grw), Memory (Gsm/Glr), Visual Processing, Auditory, Olfactory, Tactile Processing, Psychomotor and Kinesthetic abilities, etc. 
Rather than adopting the full taxonomy proposed by the CHC theory, we only retain a subset of five broad task domains with direct grounding in CHC, specifically: 1. \textbf{Quantitative knowledge} [G\emph{q}]; 2. \textbf{Fluid reasoning} [G\emph{f}] (including inductive, analogical and metaphorical reasoning); 3. \textbf{Visual processing} [G\emph{v}]; 4. \textbf{Language and Verbal Knowledge} (an aggregation of G\emph{c}/G\emph{a}/G\emph{rw} covering comprehension, lexical/grammatical competence, and reading/writing tasks); 5. \textbf{Sensory/Motor Abilities} (an ad hoc supercategory formed by collapsing \(G{v}\), \(G{o}\), \(G{h}\), \(G{p}\), and \(G{k}\)).

In line with cognitive and neuropsychological evidence emphasising the centrality of embodiment in human cognition,
we partition the domains into two macro-categories: 1. Cognitive Abilities (Quantitative, Fluid, Visual, and Language/Verbal Reasoning); 2. Sensory/Motor Abilities.

Working/short-term memory (\(Gwm\)) and long-term storage/retrieval (\(Glr/Gr\)) are not included, along with Processing speed (\(Gs\)) and decision/reaction time (\(Gt/Gps\)), since they will be captured in the \textit{Performance Match} evaluation.

Each domain has been selected for its central role in human cognitive processing. Together, they form a representative spectrum of cognitive functions that provides a diverse set of testing areas for the systems. While supported by extensive research in cognitive science \parencite{chomsky2011language, tversky1977features, kintsch1994text, gentnermarkman, holyoak2012analogy}, the list might be deemed incomplete; it is, in fact, extensible according to empirical evidence. A paramount feature of the CHC theory is that it is not static; rather, it is a dynamic model that is continuously reorganized based on research advancements. 

The capability of a model $\mathcal{M}$ in each of the $n$ task domains $\{ d_1, \dots, d_n\}$ is evaluated on a three-point ordinal scale: for each domain, let $g_j\in\{0, 0.5, 1\}$ denote the score of $\mathcal{M}$ in domain $d_j$. We assign values:

\begin{equation}
g_j =
\begin{cases}
1 & \text{if $\mathcal{M}$ covers the domain } d_j \\
0.5 & \text{if $\mathcal{M}$ partially covers the domain } d_j \\
0 & \text{otherwise}
\end{cases}
\end{equation}
The overall generality score is computed as:

\begin{equation}
G_\mathcal{M} = 0.5 \cdot \overline{g_{\text{cognitive}}} + 0.5 \cdot g_{\text{sensorimotor}}
\end{equation}
Where:

\begin{equation}
{\overline{g_{\text{cognitive}}}} = \frac{1}{n}\sum_{i=1}^ng_j 
\end{equation}
\noindent And $g_{\text{sensorimotor}}$ is the score assigned to the Sensory/Motor domain. This yields a value in the range [0,1], where higher scores indicate greater generality, i.e., the ability to operate across a wider array of cognitive functions.

\subsubsection{Performance Match}

As introduced by Lieto \parencite{lieto2021cognitive}, the Performance Match assessment should include, in addition to the evaluation of accuracy, an analysis of error patterns and task execution times. 
For each system, we can quantify the alignment between models and human subjects on selected benchmarks. The Performance Match for each model can be calculated as the sum of three different sub-metrics: 1. Percentage delta from the baseline given by human performances, 2. Error patterns, and 3. Response times. \textcolor{black}{Error patterns and execution times provide complementary behavioural constraints beyond accuracy. Models with similar overall performance may exhibit systematically different failure patterns or require different amounts and sequences of computation; accordingly, both measures have long been used in computational cognitive science \parencite{newell1994unified}},\parencite{varma2011criteria},\parencite{kunda2016error},\parencite{sartori2023language},\parencite{anderson2004integrated},\parencite{lieder2020resource}.
More formally, the aggregate Performance Match score ($PM_\mathcal{M}$) can be computed as the weighted sum: 

\begin{equation}
\label{performance_match}
{PM_\mathcal{M}} = \alpha \cdot {A_\mathcal{M}} + \beta \cdot {E_\mathcal{M}} + \gamma \cdot {T_\mathcal{M}}, \quad   
\end{equation} 

where

\begin{equation}
   \quad \alpha + \beta + \gamma= 1,\quad \alpha, \beta, \gamma \in [0, 1]
\end{equation}

\noindent
Here, $A_\mathcal{M}$ denotes accuracy, $E_\mathcal{M}$ denotes the error pattern score, and 
$T_\mathcal{M}$ denotes the matching score for execution times. Weights can be used to adjust the relative importance of each sub-dimension, depending on the evaluation goals. If no particular emphasis is needed, a uniform weighting can be applied, setting $\alpha = \beta = \gamma = \frac{1}{3}$. 

\paragraph{Accuracy} 

For $A_\mathcal{M}$, let $\mathcal{B}=\{B_1, \dots, B_n\}$ denote the set of benchmarks on which a given model is evaluated. For each benchmark $B_i \in \mathcal{B}$, we can define a $\Delta$ as difference between the model's accuracy ($A_{m,i}$) and the human baseline value: ($A_{h,i}$)

\begin{equation}
\Delta_i = A_{{m},i} - A_{{h},i}
\end{equation}

\noindent
The mean deviation from the human baseline across all $n$ benchmarks is then computed as:
\begin{equation}
\bar{\Delta} = \frac{1}{n} \sum_{i=1}^{n} \Delta_i
\end{equation}

\noindent
Then, we compute $A_\mathcal{M}$ as follows:

\begin{equation}
\label{accuracy}
{A_\mathcal{M}}=\frac{1}{1+ |\bar{\Delta}|}
\end{equation}

\noindent
This returns a maximum value of 1 if and only if the model replicates human performance on average, and decreases symmetrically as the deviation grows, whether positive or negative.

\paragraph{Error Patterns} 

To account for qualitative aspects of model performance, for each benchmark $B_i$ we define a binary Error Pattern indicator $E_i \in \{-1,+1\}$ based on whether the model replicates typical human error distributions. We assign:
\begin{equation}
E_i =
\begin{cases}
+1 & \text{if $\mathcal{M}$ replicates human error patterns} \\
-1 & \text{otherwise}
\end{cases}
\end{equation}

\noindent
This penalises systems that diverge from human failure patterns and enables a more informative scoring, especially when data on such patterns is unavailable. Then, we aggregate these across $n$ benchmarks as the mean:
\begin{equation}
\bar{E}=\frac{1}{n}\sum_{i=1}^nE_i  \quad
\end{equation}

\noindent
This results in a score $\bar{E} \in [-1,1]$. Since it is not directly comparable to other normalised scores in $[0, 1]$, we apply a transformation to map $\bar{E}$ onto the unit interval:
\begin{equation}
{E_\mathcal{M}} = \frac{\bar{E} + 1}{2} \quad \quad {E_\mathcal{M}} \in [0,1]
\end{equation}

\noindent
A score of 1 indicates reproduction of human-like errors across all benchmarks, whereas a score of 0 indicates a complete failure to replicate any human error pattern. Benchmarks for which no error data is available ($E_i = 0$) do not contribute to the score, avoiding artificial inflation or deflation of the final value.

The third component of the \textit{Performance Match} metric captures the extent to which the model replicates the response times of human cognition.
Let $t_{m,i}$ be the model's average response time on benchmark $B_i$,  and $t_{h,i}$ be the average human response time on the same benchmark. The relative temporal deviation for each benchmark is defined as:

\begin{equation}
d_i = \left| \frac{t_{m,i} - t_{h,i}}{t_{h,i}} \right|
\end{equation}

\noindent
This value expresses how far the model's response latency deviates from the human baseline in proportional terms, irrespective of whether the model is faster or slower. To convert this deviation into a similarity score bounded in the interval $[0, 1]$, we define:

\begin{equation}
T_i = \frac{1}{1 + d_i} = \frac{1}{1 + \left| \frac{t_{m,i} - t_{h,i}}{t_{h,i}} \right|}
\end{equation}

\noindent
A value of $T_i = 1$ indicates perfect alignment ($t_{m,i} = t_{h,i}$), while lower values represent increasing misalignment.

\noindent
The overall \textit{Execution Time Score} is computed as the arithmetic mean of the individual scores:

\begin{equation}
{T_\mathcal{M}} = \frac{1}{{n}} \sum_{i=1}^n T_i
\end{equation}

\noindent
In cases where timing values are only available as a proportion of human-model response time similarity (e.g., ${T_\mathcal{M}} = 0.59=59\%$), this value can be directly used:

\begin{equation}
{T_\mathcal{M}} = \text{empirical estimate} \in [0, 1].
\end{equation}

\subsubsection{Unified Metric for Cognitive Plausibility}
\label{section: unified}

To evaluate the overall cognitive plausibility of a model, we define a composite metric (${\mathcal{P}}_{\mathcal{M}}$) which integrates the three distinct dimensions analysed through the Minimal Cognitive Grid: \textit{Functional/Structural Ratio}, \textit{Generality}, and \textit{Performance Match}. The final score is computed as the  weighted sum:
\begin{equation}
{\mathcal{P}}_{\mathcal{M}} = \lambda \cdot  FSR'_{\mathcal{M}} +  \mu \cdot {G_\mathcal{M}} + \nu \cdot {PM_\mathcal{M}}\quad \text{with} \quad \lambda + \mu + \nu = 1
\end{equation}
\noindent
With $G_\mathcal{M}$ being the Generality score, and $PM_\mathcal{M}$ the Performance Match score, while $\lambda, \mu, \nu$ are non-negative weights that sum to 1 and reflect the relative importance (if any) of each dimension. \textcolor{black}{The framework assigns a higher weight to the Functional/Structural Ratio, since it is designed to be the dimension that better captures the
degree to which a model’s mechanisms align with the structural principles of human 
reasoning. Assumed that $FSR'_{\mathcal{M}}$ is the most important dimension in terms of overall cognitive plausibility for the model, we can set: $\lambda=0.5, \mu=0.25, \nu=0.25$, so that $FSR'_M$ contributes as much as the other two dimensions combined. This weighting scheme reflects the primary aim of the framework, namely, to assess whether a model can serve as a testbed for theorising about psychological phenomena in computational cognitive science. If this is the aim, adherence to the mechanisms and constraints that are taken to underlie the target cognitive phenomenon must be given priority. As argued by some authors \parencite{kaplan2011explanatory,harbecke2021methodological}, the explanatory relevance of a model requires an appropriate mapping between elements of the model and elements of the mechanism responsible for the explanandum. In this sense, the Functional/Structural Ratio is the dimension that anchors our evaluation to the explanatory profile of the model, rather than one desideratum among others. This does not imply that Generality and Performance Match are of secondary importance. On the contrary, both contribute independently to the assessment of cognitive plausibility. Alternative weighting schemes may legitimately reflect different explanatory priorities, but it remains crucial to acknowledge that high Generality or Performance Match should not compensate for the absence of the mechanisms relevant to the target cognitive phenomenon.
For this reason, we explicitly evaluate the sensitivity of the resulting cognitive plausibility assessment to alternative weighting assumptions, as reported in Table~\ref{tab:joint_weight_sensitivity}.}

Since the raw $FSR_\mathcal{M}$ values are not bounded in $[0,1]$, and lower $FSR$ values indicate stronger structural plausibility, in order to maintain consistency with the other components, we compute $FSR'_{\mathcal{M}}$ by inverting the $FSR$ value:

\begin{equation}
I(FSR_{\mathcal{M}})= \frac{1}{FSR_{\mathcal{M}}}
\end{equation}

\noindent So that higher values are guaranteed to correlate with a higher structural score. We can then use a normalisation formula to map these values into the $[0,1]$ interval:

\begin{equation}
FSR'_{\mathcal{M}} = \frac{I(FSR_{\mathcal{M}})}{1+ I(FSR_{\mathcal{M}})}
\end{equation}

\noindent
The result in $[0,1]$ can be interpreted as a structurality index: the more closely $FSR'_{\mathcal{M}}$ approaches $1$, the more strongly the model structurally aligns with the cognitive theoretical framework of reference.

The final plausibility score \( P_\mathcal{M} \) serves the purpose of offering a unified, quantitative measure of the cognitive soundness of an artificial model \( \mathcal{M} \).
The \( P_\mathcal{M} \) function yields a value in the interval \( [0, 1] \); intuitively, a system achieving a \( P_\mathcal{M} \) score close to 1 is one that (i) implements structural features grounded in cognitive theories, (ii) replicates human-like performance both in outcome and in process, and (iii) generalises across multiple cognitive domains.

Epistemologically, \( P_\mathcal{M} \) serves as an indicator to locate the target model on a continuum of explanatory  profiles through its cognitive plausibility \parencite{donvitolieto2026}. In the long term, widespread use of the \( P_\mathcal{M} \) score may contribute to a standardised typology of cognitive explanations in AI.

\subsection{Behavioural Data Replication}
\label{sec:nll_eval}

As introduced above, behavioural data replication constitutes a crucial component of this framework. We therefore conducted a series of trials to evaluate whether five of the current state-of-the-art large language models could, through a retrieval-augmented generation (RAG) instruction, \textcolor{black}{attain behavioural fit numerically close to that reported for Centaur without Psych-101-specific fine-tuning.}
The results of this first testing hypothesis are presented in the following paragraph.

More specifically, we evaluated the decision-level behavioural fit of each model using the negative log-likelihood (NLL) assigned to the observed choices, computed as follows:

\begin{equation}
\mathrm{NLL}(\theta)
=
- \sum_{t=1}^{T} \log p_\theta(a_t \mid x_t)
\end{equation}

where $x_t$ denotes the context provided to the model at time $t$, \textcolor{black}{$a_t$ is the observed human action}, $p_\theta(a_t \mid x_t)$ is the probability assigned by the model with parameters $\theta$ to that action, and $T$ is the total number of decisions in the sequence.

\textcolor{black}{Because direct token log-probabilities were not consistently available for the hosted models, choice probabilities were estimated by repeated sampling. For each of the $N=300$ human decisions ($150$ trials $\times$ two decisions), we generated $K=150$ responses at temperature $T=0$. Despite temperature = 0, repeated calls were not fully deterministic, consistently with previous evidence showing that zero-temperature decoding does not guarantee identical outputs across repeated LLM calls \parencite{ouyang2025empirical}.
Top-$p$ was left at the platform/provider default.}

\textcolor{black}{If $c_t$ is the number of generations matching the observed human action $a_t^{H}$, we estimated}

\begin{equation}
\textcolor{black}{
\hat{p}_t(a_t^{H}\mid x_t)=\frac{c_t}{K},
}
\end{equation}

\textcolor{black}{and computed}

\begin{equation}
\textcolor{black}{
\widehat{\mathrm{NLL}}_t
=
-\log \hat{p}_t(a_t^{H}\mid x_t).
}
\end{equation}

For each model, we ran $150$ trials of the two-step task, yielding $300$ total decisions per model, and computed the average NLL per decision as the primary summary metric.

\textcolor{black}{The models were instructed to return a discrete choice symbol; when additional text was produced, the corresponding legal choice symbol was parsed for scoring. No smoothing was required because no human target received zero samples. We therefore refer to this measure as a \emph{sampling-based estimate of behavioural choice NLL}. Unlike Centaur's NLL, which is derived directly from token probabilities, our estimate is Monte Carlo-based. Therefore, comparisons with Centaur should be interpreted descriptively rather than as strictly like-for-like likelihood estimates. Nonetheless, as detailed in Section 5, the provided RAG agents, instructed with a reward structure of the two-step task elicited by 150 simulations via the  Spaceship game also used to fine-tune Centaur (\cite{binz2025foundation}, p. 1005), shows that our setup reaches performances comparable to Centaur without requiring any claim of “cognitive” or “human-like” mechanisms implied in the provided output. Notably, while Centaur was fine-tuned with the reward structures of the two-step task (extracted by the same Spaceship game mentioned above, but not publicly available) here we just presented to our RAGs agents a possible configuration of the reward structure of the task showing that, without fine-tuning, we match the performances of the Centaur model.}

\subsection{\textcolor{black}{fMRI-based Methodological Comparison}} 

A final component of LAPITHS is constituted by \textcolor{black}{the use of available neural data to assess how informative neural-similarity measures are when applied to language-model-based systems.}\footnote{\textcolor{black}{Neural activity may be viewed as a further form of behaviour, and its relevance to cognition depends on substantive assumptions about the relationship between neural processes and cognitive functions. Our use of neural evidence follows the standard methodological commitments of cognitive neuroscience, where cognition is taken to arise from underlying neural and bodily processes. Whether this assumption is ultimately correct is an empirical matter. The present analysis adopts this evidential role of neural data without advancing any additional metaphysical claims about the nature of cognition.}}

\textcolor{black}{Our analysis should not be interpreted as a strict replication of Centaur's neural-decoding experiment. Rather, using the same Two-Step Task fMRI dataset, we test whether high ROI-level similarity can also arise from a substantially different RAG+LLM procedure that does not decode internal model representations.}

In order to better explain and introduce this element of comparison it is necessary to introduce the arguments provided in Centaur. We provide an overview below. 

\subsubsection{fMRI-based representation in Centaur}

A central question raised by the Centaur framework concerns whether fine-tuning can genuinely bring the internal representations of an LLM closer to those observed in human cognition. This issue underlies one of the most significant claims advanced by the authors. Centaur proposes that model-derived internal representations can be aligned with human neural data by mapping model states to fMRI responses recorded during decision-making tasks. According to this view, if Centaur’s internal representations correspond to those observed in human participants during or after task performance, or if the model can predict neural activity in datasets on which it was not trained, then the model would not merely reproduce behavioural outputs but would also approximate the representational processes underlying them.
To support this claim, Binz and colleagues attempted to predict human neural activity using Centaur’s internal representations. In their first analysis, whole-brain fMRI measurements obtained from participants performing the Two-Step Task (TST) are used as the target signal. Neural recordings are extracted both prior to each decision and following reward feedback. These signals are aggregated within each brain region and regressed against Centaur’s internal representations, with the procedure repeated independently for each participant and each region \parencite[p.~5]{binz2025foundation}. The neural dataset employed in this analysis originates from a study  \parencite{feher2023rethinking}, which investigated learning mechanisms in sequential decision-making tasks. Participants performed a two-stage decision paradigm presented in both abstract and narrative variants. In the Centaur study, however, this dataset is primarily used as a source of neural recordings rather than as a theoretical object of analysis. To evaluate predictive performance, Binz and colleagues use Centaur’s internal representations as well as those extracted from the base model LLaMA in order to assess the effect of fine-tuning. Prediction is performed through a regularised linear regression model. Each participant and each brain region constitutes a separate dataset, which is divided into training and testing subsets, typically using two-thirds of the data for training and one-third for evaluation. Model representations are extracted from the residual stream of the transformer architecture both before decisions and after feedback. These representations are subsequently reduced through principal component analysis (PCA) so as to retain the principal directions explaining approximately 95\% of the variance. The resulting components are then used as predictors of neural activity within regions of interest defined according to the Harvard–Oxford atlas.
In transformer models, internal representations consist of high-dimensional vectors encoding the informational state of the network at each layer. The residual stream functions as a continuous pathway through which representations accumulate across layers via residual connections that combine transformed outputs with their corresponding inputs. While technically coherent, this procedure relies on a controversial assumption: that the internal representations of an LLM can meaningfully be compared with human neural representations \parencite{quattrociocchi2025epistemological, quattrociocchi2026statistical}. As discussed earlier, and as we show in the following, models such as Centaur exhibit at most functional similarities with human cognition rather than structural correspondences. The analysis therefore implicitly adopts a strongly functionalist position according to which functional similarity is sufficient to justify comparisons between systems with fundamentally different architectures. 
The methodological framework underlying this approach is derived from the comparative benchmarking strategy proposed by \parencite{schrimpf2021neural}. Their work investigated whether artificial neural networks trained on language tasks could predict neural responses recorded during language processing. Importantly, the authors themselves emphasised that such models should be regarded only as simplified approximations of certain aspects of neural processing rather than faithful reconstructions of brain mechanisms. Nevertheless, Schrimpf et al. showed that transformer-based models could account for a substantial proportion of the explainable variance in neural responses to sentences across multiple datasets and imaging modalities. Models that performed well on next-word prediction tasks also tended to predict neural and behavioural responses more accurately. GPT-2, in particular, exhibited strong performance across both linguistic tasks and neural predictivity metrics. These results were interpreted cautiously as evidence that contemporary language models might capture some statistical regularities relevant to human language processing.
However, the ability to reproduce behavioural or neural patterns does not necessarily imply that the underlying cognitive mechanisms are shared. A model may predict human responses accurately simply by capturing statistical regularities present in linguistic data, without implementing processes analogous to those occurring in the brain or at the cognitive level \parencite{quattrociocchi2026statistical}. The Centaur study adopts essentially the same benchmarking strategy. Empirical results reported in the study further complicate the interpretation. In several cases the difference in predictive performance between Centaur and LLaMA diminishes in deeper layers of the model, with the predictive scores of the two systems converging as the representational hierarchy progresses. If fine-tuning produces only marginal differences in layers likely to encode more abstract representations, the claim that Centaur’s internal representations are substantially more aligned with human neural activity becomes difficult to sustain. More broadly, comparative benchmarking may indeed provide a useful method for identifying functional correspondences between artificial and biological systems. Extending such findings to claims about alignment between internal representations is, however, considerably more problematic. Improvements in predictive accuracy following fine-tuning may simply reflect the model’s adaptation to the dataset used during training. Given current machine-learning practices, it is expected that a model trained on a specific dataset will perform better on tasks derived from that dataset.
For these reasons, the conclusion proposed by Binz et al.—namely that Centaur’s representations outperform those of LLaMA in predicting neural activity and therefore become aligned with human neural representations—appears insufficiently supported. Fine-tuning clearly improves performance on specific predictive tasks, but this improvement does not by itself establish representational equivalence. A further methodological concern relates to the limited transparency surrounding the pretraining data used for LLaMA models. Because the precise composition of the training corpus for LLaMA 3.1 70B has not been publicly disclosed, it remains possible that the base model has already encountered materials related to neuroscience, including descriptions of neural imaging data. If so, part of the predictive capacity attributed to Centaur may already have been present in the underlying model. 
Taken together, these findings suggest that the analytical procedure primarily demonstrates the effectiveness of fine-tuning in improving task-specific performance rather than establishing a meaningful alignment between Centaur’s internal representations and those of the human brain. \textcolor{black}{For this reason, rather than attempting a strict replication of Centaur's hidden-state decoding procedure \parencite{binz2025foundation}, we conducted a complementary methodological test asking whether high ROI-level correlation can also emerge from a more weakly constrained RAG+LLM generation procedure.}


\section{LAPITHS Evaluation on the Two-Step Task}\label{sec:results}
The above described elements of LAPITHS are adopted below by focusing on a specific but relevant task included in Psych-101: the Two-Step Task (TST). It represents an example of a sequential decision-making paradigm in the context of reinforcement learning (RL) research. It requires participants to make two consecutive choices across two stages. A first-stage choice probabilistically leads to one of two second-stage states (with common transitions occurring with probability 0.7 and rare transitions with probability 0.3). At the second stage, participants make another choice that may yield a reward according to probabilities that slowly drift over time via a random-walk process. 
Since rewards are provided at the second stage, optimal performance needs learning which second-stage options are most rewarding and selecting first-stage actions that increase the likelihood of reaching those states. Behavioural and neuroscientific evidence \parencite{daw2011model} shows that human behaviour in this task reflects the interaction between two classes of RL strategies: \textit{model-free} and \textit{model-based}. 

In the context of Centaur, TST is used by the authors \parencite{binz2025foundation} as a central test to demonstrate the generalization capabilities of their model. In particular, the authors report that Centaur is able to generalize effectively when the cover story of the task is modified.

In its original formulation, indeed, the TST is presented through a spacecraft narrative \parencite{daw2011model}. In particular: the participant is required to maximise their reward by making sequential choices across two stages. In the first stage, the player selects between two rockets, each of which leads to either Planet A or Planet B. Upon arriving at the selected planet, the participant faces a second decision between two aliens, each presenting a box that may contain a treasure—denoted X or Y. After the final selection, the participant is informed whether a treasure has been obtained. As noted above, the objective of the task is to maximise the total reward obtained across trials. The authors argue that Centaur can successfully generalise the underlying structure of the task when the narrative context is altered \parencite[p.~4]{binz2025foundation}. Specifically, they replace the spacecraft scenario with an alternative magic carpet cover story.\footnote {In the “magic carpet” story, the participant is asked to imagine travelling on a flying carpet between different locations. At each trial, they first choose a destination, which probabilistically leads to one of two possible intermediate places. From there, they make a second choice that may yield a reward (e.g., finding treasure), with reward probabilities that change over time. The logical structure of the task is identical to the “spaceship” paradigm, differing only in its narrative presentation.}  Although the narrative elements and choice options are modified accordingly, the underlying decision structure remains identical. According to their results, Centaur is capable of identifying and solving the task despite these changes in presentation, outperforming several baseline models. 

In Centaur, however, the relevance of TST extends well beyond its role as a test of generalisation (and thus the reason why we selected it). In their paper, the authors further claim that their model develops internal representations that become increasingly aligned with \textcolor{black}{human neural activity}. The TST experiment is presented, together with the next-word prediction task, as a primary piece of evidence supporting this claim (ibid., p. 5). A critical examination of this aspect of their argument will be addressed in Section \ref{sec:fmri_test}. First, we focus on the results of the analysis based on the application of the Minimal Cognitive Grid and, in the next paragraphs, we provide counter arguments concerning the behavioural data on the task as well as the claims that the authors make about their fMRI representation comparison. \textcolor{black}{As detailed below, our behavioural and neural analyses are intended as methodological comparisons rather than strict replications of Centaur's evaluation procedures.}

\subsection{MCG Analysis}
\label{sec:MCG}
We computed Centaur's ${\mathcal{P}}_{\mathcal{M}}$ score with respect to the data regarding the two-step task.  
Before proceeding to the evaluation of Centaur along the three dimensions of 
the Minimal Cognitive Grid, it is necessary to establish the cognitive constraints against which the model will be assessed. These constraints are derived from the requirements of the two-step task as executed by human participants, and constitute the reference framework for the Functional/Structural Ratio analysis.

For an agent to perform the task, its architecture must implement at least three components: 
(1) an incremental learning mechanism driven by feedback, characteristic of model-free reinforcement learning \parencite{degris2012model, akam2015simple}. Under the canonical reinforcement-learning interpretation of the two-step task, feedback is assumed to modify value estimates incrementally across trials through reward-prediction-error signals.
Formally, value updating can be expressed as:
\begin{equation}
Q_{t+1}(s,a)=Q_t(s,a)+\alpha \delta_t
\end{equation}
where \(Q(s,a)\) denotes the estimated value of action \(a\) in state \(s\), \(\alpha\) is the learning rate, and \(\delta_t\) is the reward prediction error at time \(t\). These conditions are satisfied when feedback received at time $t$ produces an immediate, local update of the estimates guiding subsequent choices, without requiring the accumulation of multiple experiences \parencite{schultz2016dopamine, gershman2017reinforcement}. 
This requirement concerns the online updating of the model's parameters during the execution of the task. It should be kept distinct from the model’s training regime, which in modern systems is typically based on batch gradient optimization and need not correspond to the biological mechanisms of synaptic plasticity \parencite{lillicrap2020backpropagation};
(2) an evaluation mechanism that exploits the transition structure of the environment, characteristic of model-based reinforcement learning. In fact, solving the two-step task requires the ability to evaluate first-stage actions using an internal representation of the environment's transition structure [model-based RL, \parencite{m2023model}].
Within a model-based strategy, the value of an action is computed by combining transition probabilities with the values of successor states. Formally, the value of a first-stage action can be expressed as:
\begin{equation}
\label{model_based}
Q(s,a)=\sum_{s'} T(s,a,s')V(s')
\end{equation}
where \(T(s,a,s')\) denotes the probability that action \(a\) in state \(s\) leads to successor state \(s'\), and \(V(s')\) denotes the value of the successor state \parencite{daw2011model}.
This mechanism allows the system to plan and evaluate the consequences of its choices by exploiting a model of the causal structure of the task. In the two-step task, this mechanism gives rise to the characteristic behaviour of model-based strategies, namely switching the first-stage choice when reward follows a rare transition;
(3) a memory system that allows relevant information to be maintained between the first decision and the final outcome. Since in the two-step task rewards are obtained only after a sequence of events -- including the initial choice, the transition to a second-stage state, and a second decision -- in order for the system to update the values of first-stage actions, it must retain relevant information during the interval separating the action from the outcome. In humans, this process relies on working memory, localised in the dorsolateral prefrontal cortex and frontoparietal networks \parencite{curtis2003persistent}. Working memory capacity is limited to approximately three to four items \parencite{cowan2001magical} and is subject to decay when information is not actively maintained \parencite{baddeley2003working}. Consistent with this, overloading working memory reduces the influence of model-based strategies and increases the contribution of model-free control in the two-step task \parencite{otto2013principles}. The ``memory'' constraint is thus divided into two sub-components, the first for information persistence and the second for memory capacity limit, each weighted by 1/6. 
These components define the minimal set of structural constraints that a system must satisfy in order to solve the task according to the principles identified in the scientific literature on human RL. 
In what follows, we evaluate Centaur against each of the constraints in order to compute its Functional/Structural Ratio score, starting with (1): Centaur does not satisfy the requirement of incremental, local, and immediate feedback-driven learning. It was trained via QLoRA fine-tuning \parencite{dettmers2023qlora}, which relies on backpropagation to minimise a cross-entropy loss over batches of trials, and is neither local nor immediate. Furthermore, model parameters are frozen at inference time, meaning that no weight update of any kind occurs during the execution of the task. Accordingly, we assign $s_1 = 0$.

(2) Centaur satisfies the requirement that relevant information be actively maintained between the first-stage choice and the delivery of feedback. The model receives the full trial-by-trial history of the session as a natural language prompt, and its self-attention mechanism allows it to retrieve information within the context window. Accordingly, we assign $s_{2a} = 1$. It should be noted, however, that Centaur's implementation of this constraint diverges considerably from the biological profile of human working memory: the context window imposes no capacity limit comparable to the approximately three to four items retained by humans \parencite{cowan2001magical}, and no decay or interference effects are present ($s_{2b} = 0$).

(3) Centaur does not implement an explicit mechanism for model-based evaluation of first-stage actions. The model-based strategy requires a system to compute $Q(s,a) = \sum_{s'} T(s,a,s') V(s')$, that is, to 
explicitly combine a representation of transition probabilities with the values of successor states in order to derive the value of first-stage actions. Centaur 
approximates action probabilities via a single forward pass through the transformer architecture, with no explicit representation of $T(s,a,s')$. Behaviour consistent with model-based strategies emerges from the forward pass, but such emergence constitutes a functional implementation of the constraint. Accordingly, we assign $s_3 = 0$.

The resulting scores are summarised in Table~\ref{tab:centaur_fsr}.
Raw functional/structural ratio values $F/S + \epsilon>1$ indicate that the model relies on behaviouristic output resemblance, and deviates significantly from the cognitive 
constraints postulated by the reference theories of human reinforcement learning.

The analysis of Generality follows the structure presented in the previous section. Out of the five selected task domains in our framework, Centaur covers Quantitative Knowledge, Fluid Reasoning and Language/Verbal Knowledge, while lacking abilities in the Visual/Spatial and Sensory/Motor domains. 
With respect to Quantitative Knowledge, the model inherits the mathematical reasoning capabilities of Llama 3.1 70B, and the fine-tuning procedure on Psych-101 does not degrade performances on quantitative benchmarks, as verified by the authors $(g_q = 1)$.

As for Fluid Reasoning, Centaur shows coverage of inductive reasoning, abstract reasoning and decision-making under uncertainty across a broad range of experimental paradigms included in Psych-101  $(g_f = 1)$. 
Same for Language and Verbal Knowledge, which is the primary domain of LLMs in general $(g_l = 1)$. 
On the other hand, Centaur is a unimodal text-based model, which possesses no mechanisms for processing or reasoning about visual or spatial inputs $(g_v = 0)$ and, like all disembodied systems, lacks perceptual interfaces and physical effectors, and cannot instantiate any form of perception-action dynamics with the environment $(g_{sm} = 0)$.

Finally, we evaluate Centaur on each component of the Performance Match dimension, with respect to the two-step task data reported by \parencite{binz2025foundation}: (1) \textcolor{black}{Accuracy can be computed using negative log-likelihood (NLL), which measures how well the model predicts the observed choices of human participants, under the assumption that NLL = 0 corresponds to perfect prediction of every human choice. The two-step task does not provide a direct human-performance baseline analogous to the proportion of correct responses in a task with objectively correct answers: participants choose between alternatives whose reward probabilities change over time, and the task is designed to identify the learning and decision processes generating those choices.}

\textcolor{black}{To estimate the closest available empirical proxy for the human baseline against which assessing the model's performance, we used the human-derived predictive Reinforcement Learning model provided in the official Centaur repository.}\footnote{\textcolor{black}{The \textit{DualSystemsModel} is a simplified version of the hybrid model-based/model-free reinforcement-learning architecture introduced for the two-step task by \parencite{daw2011model}. It maintains model-free action values updated trial by trial through SARSA($\lambda$). Given first-stage action $a_{1,t}$, second-stage state $s_{2,t}$, second-stage action $a_{2,t}$, and reward $r_t$, it computes $\delta_{1,t}=Q_t^{\mathrm{MF}}(s_{2,t},a_{2,t})-Q_t^{\mathrm{MF}}(s_1,a_{1,t})$ and $\delta_{2,t}=r_t-Q_t^{\mathrm{MF}}(s_{2,t},a_{2,t}),$ and applies the updates $Q^{\mathrm{MF}}(s_1,a_{1,t})\leftarrow Q^{\mathrm{MF}}(s_1,a_{1,t})+\alpha\delta_{1,t}+\lambda\alpha\delta_{2,t}$ and $Q^{\mathrm{MF}}(s_{2,t},a_{2,t})\leftarrow Q^{\mathrm{MF}}(s_{2,t},a_{2,t})+\alpha\delta_{2,t}.$ Model-based first-stage values are computed from the task’s fixed $T=.7/.3$ transition matrix through $Q_t^{\mathrm{MB}}(a_1)=\sum_{s_2}T(s_2\mid a_1)\max_{a_2}Q_t^{\mathrm{MF}}(s_2,a_2).$ First-stage choices depend on $Q_t^{\mathrm{net}}(a_1)=\tau Q_t^{\mathrm{MB}}(a_1)+(1-\tau)Q_t^{\mathrm{MF}}(s_1,a_1)$ plus a stickiness term $\rho\mathbb{I}(a_1=a_{1,t-1}),$ whereas second-stage choices depend on the corresponding model-free values; both are converted into choice probabilities by a softmax with fitted scale parameter $\beta$. The five shared parameters $\theta=(\tau,\alpha,\lambda,\rho,\beta)$ are estimated from the \parencite{feherdasilva2020humans} dataset by minimizing the mean cross-entropy, equivalent to choice NLL, over both decision stages. The repository performs ten-fold participant-level cross-validation: in each fold, a new model is fitted for 1,000 optimization steps to all trials from nine participant subsets and evaluated, without further fitting, on the participants in the remaining subset. Averaging the ten held-out NLL values yields $NLL_H=0.58$.}}
\textcolor{black}{Its predictive performance is evaluated on held-out participants using ten-fold participant-level cross-validation, yielding a mean NLL of 0.5882. The value quantifies how well a TST-specific cognitive model whose parameters are estimated from human behavioural data predicts the choices of unseen human participants. Centaur achieves a mean NLL of $0.51$ on the same prediction task of human responses. We therefore obtain $
|\Delta| = |0.51 - 0.58| = 0.07,$
which, applying Eq.~\eqref{accuracy},
yields $A_M \approx 0.93$.}

(2) Assessment of error patterns. In the two-step task, the optimal strategy is purely model-based: an agent that computes equation \eqref{model_based} at each trial and selects actions accordingly maximises expected reward by exploiting the known 
transition structure. Any deviation toward a model-free strategy amounts to a suboptimal response. Under this 
definition, the human error profile in the two-step task is characterised by the distribution of model-basedness across participants, reflecting how frequently, and to what degree, human participants rely on the suboptimal model-free strategy rather than the optimal model-based one, including the full range of individual 
variation from purely model-free to purely model-based behaviour.

The authors report data from open-loop performances of Centaur on the two-step task (Figure 2c), showing that the model reproduces the bimodal distribution of the model-basedness parameter observed in human participants, with comparable density across the model-free, mixed, and model-based regions of the parameter space. The correspondence between the two distributions indicates that Centaur replicates the structure of human suboptimal behaviour. Accordingly, we assign $E_M = 1$.

(3) For response time, the authors propose an indirect measure of the alignment between the model and human subjects in Psych-101 based on Hick's law \parencite{hick1952rate, hyman1953stimulus}, according to which response time increases with the uncertainty of the choice. The authors show that Centaur's response entropies account for a substantial proportion of the variance in human response times (conditional $R^2 = 0.87$) -- outperforming both Llama (0.75) and domain-specific cognitive models (0.77) --, and interpret this result as evidence that the model captures something beyond choice distributions alone.
This interpretation is controversial, since Centaur was fine-tuned to predict human choices, and its output distributions approximate the empirical distribution of human responses by construction. Response entropy is a function of these output probabilities, carrying no information beyond what is already contained in the choice distributions themselves. The correlation between Centaur's entropy and human response times is unsurprising, given that it is a direct consequence of the training objective: a model that perfectly approximated human choice probabilities would by definition also perfectly predict response time variance under Hick's law, without providing any independent cognitive information about processing time. The high $R^2$ reported by the authors is better understood as a restatement of Centaur's goodness-of-fit to human choice data. More fundamentally, the measure does not satisfy the requirements of the $T_M$ component of the MCG. Hick's law is a regularity of human cognition, but it does not apply to a computational model whose inference time is determined by hardware and computational load, independently of the difficulty of individual trials as perceived by human subjects.
Applying formula \eqref{performance_match}, we obtain $PM_\mathcal{M}\approx0.96$. The score does not consider execution time data and is computed as the sum of Accuracy and Error Patterns, each weighted by 0.5.

Tables~\ref{tab:centaur_fsr}--\ref{tab:centaur_cog_plausibility_overall} summarise Centaur's scores across the three dimensions of the 
Minimal Cognitive Grid. The overall cognitive plausibility score $\mathcal{P}_M$ is computed as the weighted sum of the normalised structurality index $FSR'_M$, the 
generality score $G_M$, and the performance match score $PM_M$, following the weighting scheme proposed in section \ref{section: unified}. The overall cognitive plausibility score is therefore $\mathcal{P}_M = 0.42$.
Centaur is characterised by a strong performance 
match, which reflects the model's ability to reproduce the behavioural patterns of humans in the two-step task, and by a high generality score, consistent with its design as a domain-general foundation model. The $FSR'_M$ index, by contrast, is low, suggesting that the mechanisms underlying the model's behaviour diverge from the constraints postulated by the reference theories of human reinforcement learning. This dissociation may represent the primary limitation of Centaur as a cognitive model of human decision-making in the two-step task.

\begin{table*}[!ht]
\centering
\caption{Centaur's Functional/StructuralRatio $(FSR_M)$. Each constraint ($C_1\dots C_4$) is a requirement for cognitively plausible reinforcement learning. The model satisfies only the memory persistence constraint ($C_4 = 1$), while the remaining are implemented functionally. The resulting scores ($S_M = 0.17$, $F_M = 0.83$) yield a high functional-to-structural ratio ($F/S + \varepsilon = 4.72$), indicating a weak alignment with human cognitive processes. $FSR_M = 0.17$ is the normalized score.}
\label{tab:centaur_fsr}

\setlength{\tabcolsep}{6pt}

\begin{tabular*}{\textwidth}{@{\extracolsep\fill}lcccccccc@{}}
\toprule
\textbf{Constraint} & \textbf{C}$_1$ & \textbf{C}$_2$ & \textbf{C}$_3$ & \textbf{C}$_4$ & $\mathbf{S_M}$ & $\mathbf{F_M}$ & $\mathbf{F/S + \varepsilon}$ & $\mathbf{FSR_M}$ \\
\midrule
\textbf{Weight}     & 0.33 & 0.33 & 0.17 & 0.17 & 0.17 & 0.83 & 4.72 & 0.17 \\
\textbf{Structural} & 0.00 & 0.00 & 0.00 & 1.00 & /    & /    & /    & /    \\
\textbf{Functional} & 1.00 & 1.00 & 1.00 & 0.00 & /    & /    & /    & /    \\
\bottomrule
\end{tabular*}
\end{table*}

\subsubsection{Sensitivity Analysis}
\label{sec:sensitivity}

\textcolor{black}{To determine whether the conclusions drawn from the MCG analysis remain consistent when our weighting assumptions vary within reasonable ranges, we performed two sensitivity analyses by perturbing both the weights assigned to the cognitive constraints and those assigned to the individual MCG dimensions around their baseline configurations. Each weight was  perturbed by $\pm 30\%$ of its value, while the remaining weights were proportionally rescaled so that their relative proportions were preserved and the their sum remained equal to one. The two analyses were then crossed. Table \ref{tab:joint_weight_sensitivity} reports, for each constraint-weight perturbation, the corresponding value of $FSR'_M$ and the overall
cognitive plausibility score obtained under the baseline dimension
weights and under each $\pm30\%$ perturbation of $\lambda$, $\mu$, or
$\nu$.}

\textcolor{black}{The results show that the assessment remains stable throughout the examined perturbation range: the "structurality" score varies between $FSR'_M=0.13$ and $FSR'_M=0.22$. In terms of the underlying raw Functional/Structural Ratio, these values correspond to $FSR_M=6.97$ and $FSR_M=3.46$, well above $FSR_M=1$ (which, in the present formulation, marks the conventional boundary between functional and structural models). The largest improvement in structurality is obtained by increasing $w_4$, as expected, since $C_4$ is the only constraint satisfied by Centaur. Conversely, reducing $w_4$ gives the lowest structurality score. When perturbations at the two weighting levels are combined, the full observed range is $0.32 \leq P_M \leq 0.51$, with the minimum occurring when $w_4$ is decreased by $30\%$ while $\lambda$ is increased by $30\%$, and the maximum occurring under the converse configuration ($w_4$ increased by $30\%$ and $\lambda$ decreased by $30\%$).}

\begin{table}[t]
\centering

\begingroup
\color{black}
\arrayrulecolor{black}

\caption{Sensitivity analysis of Centaur's cognitive plausibility to variations in the
cognitive constraint weights and in the MCG dimensions weights. For each perturbation, one weight
is varied by $\pm30\%$ while the remaining weights are proportionally renormalised to lie in the unit interval.}
\label{tab:joint_weight_sensitivity}

\small
\setlength{\tabcolsep}{3pt}
\renewcommand{\arraystretch}{1.08}

\begin{tabular*}{\textwidth}{
@{\extracolsep{\fill}}
l c c c c c c c c
@{}}
\toprule

\multirow{2}{*}[-0.2em]{\shortstack[c]{\textbf{Constraint}\\\textbf{perturbation}}} &
\multirow{2}{*}[-0.2em]{{\bfseries\boldmath $FSR'_M$}} &
\multicolumn{7}{c}{\textbf{Overall cognitive plausibility ($P_M$)}} \\

\cmidrule(lr){3-9}

&
&
\textbf{Base} &
\shortstack{\boldmath$\lambda$\\$+30\%$} &
\shortstack{\boldmath$\lambda$\\$-30\%$} &
\shortstack{\boldmath$\mu$\\$+30\%$} &
\shortstack{\boldmath$\mu$\\$-30\%$} &
\shortstack{\boldmath$\nu$\\$+30\%$} &
\shortstack{\boldmath$\nu$\\$-30\%$} \\

\midrule

Baseline     & 0.17 & 0.42 & 0.35 & 0.50 & 0.42 & 0.43 & 0.48 & 0.37 \\

$C_1 +30\%$ & 0.15 & 0.41 & 0.33 & 0.49 & 0.41 & 0.41 & 0.46 & 0.35 \\
$C_1 -30\%$ & 0.20 & 0.43 & 0.36 & 0.50 & 0.43 & 0.44 & 0.49 & 0.38 \\

$C_2 +30\%$ & 0.15 & 0.41 & 0.33 & 0.49 & 0.41 & 0.41 & 0.46 & 0.35 \\
$C_2 -30\%$ & 0.20 & 0.43 & 0.36 & 0.50 & 0.43 & 0.44 & 0.49 & 0.38 \\

$C_3 +30\%$ & 0.17 & 0.42 & 0.34 & 0.49 & 0.41 & 0.42 & 0.47 & 0.36 \\
$C_3 -30\%$ & 0.18 & 0.43 & 0.35 & 0.50 & 0.42 & 0.43 & 0.48 & 0.37 \\

\addlinespace[2pt]

$C_4 +30\%$ &
\textbf{0.22}\,{\tiny (max)} &
0.45 &
0.38 &
\textbf{0.51}\,{\tiny (max)} &
0.44 &
0.45 &
0.50 &
0.39 \\

$C_4 -30\%$ &
\textbf{0.13}\,{\tiny (min)} &
0.40 &
\textbf{0.32}\,{\tiny (min)} &
0.48 &
0.39 &
0.40 &
0.45 &
0.34 \\

\bottomrule
\end{tabular*}
\arrayrulecolor{black}
\endgroup
\end{table}

\begin{table*}[!t]
\centering
\caption{Generality evaluation across five domains derived from the CHC model. Centaur covers Quantitative Knowledge, Fluid Reasoning, and Language/Verbal abilities. It lacks capabilities in Visual Processing and in the Sensory/Motor domain. Generality score ($G_M = 0.37$) is computed by assigning equal weight (0.5) to (i) the average score across the four cognitive domains and (ii) the Sensory/Motor domain.}
\label{tab:centaur_generality}

\small
\setlength{\tabcolsep}{4pt}

\begin{tabular*}{\textwidth}{@{\extracolsep{\fill}}lcccccc@{}}
\toprule
\textbf{Model} 
& \makecell{\textbf{Quant.}\\\textbf{Know.}} 
& \makecell{\textbf{Fluid}\\\textbf{Reas.}} 
& \makecell{\textbf{Vis.}\\\textbf{Proc.}} 
& \makecell{\textbf{Lang. \&}\\\textbf{Verb.}} 
& \makecell{\textbf{Sens./}\\\textbf{Mot.}} 
& $\mathbf{G_M}$ \\
\midrule
Centaur & 1.00 & 1.00 & 0.00 & 1.00 & 0.00 & 0.37 \\
\bottomrule
\end{tabular*}
\end{table*}

\begin{table*}[!t]
\centering
\caption{Performance Match ($PM_M = 0.83$) evaluation on the Two-Step Task (TST). Accuracy is computed from the deviation in NLL from the human baseline. Centaur reproduces human-like error distributions in terms of model-based versus model-free behaviour ($E_M = 1$). Response time score is not available.}
\label{tab:centaur_pm}

\small
\setlength{\tabcolsep}{4pt}

\begin{tabular*}{\textwidth}{@{\extracolsep{\fill}}lccccccc@{}}
\toprule
\textbf{Model} 
& \textbf{Task} 
& \makecell{\textbf{Human}\\\textbf{baseline}} 
& \makecell{$\mathbf{|\Delta|}$\\\textbf{NLL}} 
& \textbf{Accuracy} 
& \makecell{\textbf{Error}\\\textbf{Patt.}} 
& \makecell{\textbf{Resp.}\\\textbf{Times}} 
& $\mathbf{PM_M}$ \\
\midrule
Centaur 
& \makecell{Two-Step\\Task} 
& 0.58 
& 0.07 
& 0.93 
& 1.00 
& / 
& 0.96 \\
\bottomrule
\end{tabular*}
\end{table*}

\begin{table*}[!t]
\centering
\caption{Overall cognitive plausibility score on the MCG. The final score ($P_M = 0.42$) combines the normalized structurality index ($FSR_M = 0.17$), generality ($G_M = 0.37$), and performance match ($PM_M = 0.96$), with weights $\lambda = 0.5$, $\mu = 0.25$, and $\nu = 0.25$.}
\label{tab:centaur_cog_plausibility_overall}

\setlength{\tabcolsep}{6pt}

\begin{tabular*}{\textwidth}{@{\extracolsep\fill}lcccc@{}}
\toprule
\textbf{Model} & $\mathbf{FSR}_M$ & $\mathbf{G_M}$ & $\mathbf{PM}_M$ & $\mathbf{P}_M$ \\
\midrule
Centaur & 0.17 & 0.37 & 0.96 & 0.42 \\
\bottomrule
\end{tabular*}
\end{table*}


\subsection{Behavioural Data Analysis}
In this section we detail the element of the behavioural data replication of LAPITHS over Centaur on the Two-Step Task (TST). 

\subsubsection{\textcolor{black}{Task 1: Sampling-based NLL comparison}}

In this case, our initial objective was to test whether comparable results could be obtained using standard large language models not specifically trained on the Psych 101 dataset but equipped with RAG (Retrieval-Augmented Generation). In other words, we sought to determine whether the reported generalisation performance might be reproduced without requiring the specialised fine-tuning procedure used to construct Centaur. 

Our RAG systems (for their deployment we employed the chatbase platform supporting the selection of multiple LLMs as basis) were equipped only with the information about the reward scheme of the task.\footnote{An example of the schema, provided in json file, is available in the supplementary materials. \textcolor{black}{The RAGs configuration of our pipeline is the following: temperature = 0, with top\_p = 1 (nucleus-sampling parameter),  K = 150 repeated responses per decision, and 150 two-step trials per model. The RAGs used OpenAI’s text-embedding-3-large (3072-dimensional vectors), structure-aware JSON chunking, Supabase/pgvector as vector store, and cosine-similarity retrieval with Top\_K = 1. The retrieved context was then provided to different generative LLMs: Llama 4 Maverick, ChatGPT 4o and 5.1, Gemini 2.5 Pro, and DeepSeek R1.}}  

In such RAGs, the task is presented with the Spacecraft narrative but, like Centaur, it can handle-without any explicit training-interchangeable narratives of the tasks like the Magic Carpet (i.e., to obtain an answer, the user can rely on any narrative based on the underlying structure of the task). The interface of our multi-RAG system is available and testable at the following link: \url{https://www.ciitlab.org/two-step-task.html}. The entire instruction tuning used for its deployment is also provided in the supplementary material.

This first analysis compares the average NLL per decision across our evaluated models under a fixed two-step task scheme and scoring protocol. Figure~\ref{fig:nll_ours} reports the mean NLL per decision for our models only, providing a focused view of relative performance within our pipeline. Lower values indicate that the model assigns higher probability to the observed actions, hence better agreement with the empirical decision distribution. \textcolor{black}{As specified in Section~\ref{sec:nll_eval}, these values are sampling-based estimates rather than direct token-log-probability NLLs.}

\subsubsection{\textcolor{black}{Task 2: Approximate statistical comparison vs baselines}}

The second analysis positions our models against published baselines reported in the Centaur study, namely Centaur, Llama~3.1, and a cognitive model.\footnote{\textcolor{black}{The cognitive model  used by \parencite{binz2025foundation} for the Magic Carpet evaluation is the hybrid RL model combining model-based and model-free control introduced by \parencite{daw2011model}. In the out-of-distribution evaluation protocol used by Binz et al., the model was not fitted directly to the Magic Carpet participants, but its parameters were estimated on the most similar experiment available in the training set, namely a Two-Step Task employing the standard spaceship cover story, and the resulting parameterisation was then used to predict responses in the unseen Magic Carpet experiment.}} Figure~\ref{fig:nll_vs_baselines} shows the same mean NLL per decision values augmented with these reference systems. 

\textcolor{black}{As a supplementary inferential comparison,} we conducted Welch two-sample $t$-tests comparing each model's decision-level NLL samples to each baseline distribution. Since the baseline paper reports confidence intervals rather than raw per-decision samples, we estimated baseline standard deviations by inverting the reported CI$_{95}$ under the assumption of $n=300$ decisions. \textcolor{black}{The resulting tests are therefore approximate and should not be interpreted as strict like-for-like cross-study significance tests.} All baseline comparisons in this section use the Welch formulation and the inferred baseline variance.

\subsubsection{NLL results}
\label{sec:nll_results}

\begin{figure*}[!t]
\centerline{\includegraphics[width=\linewidth]{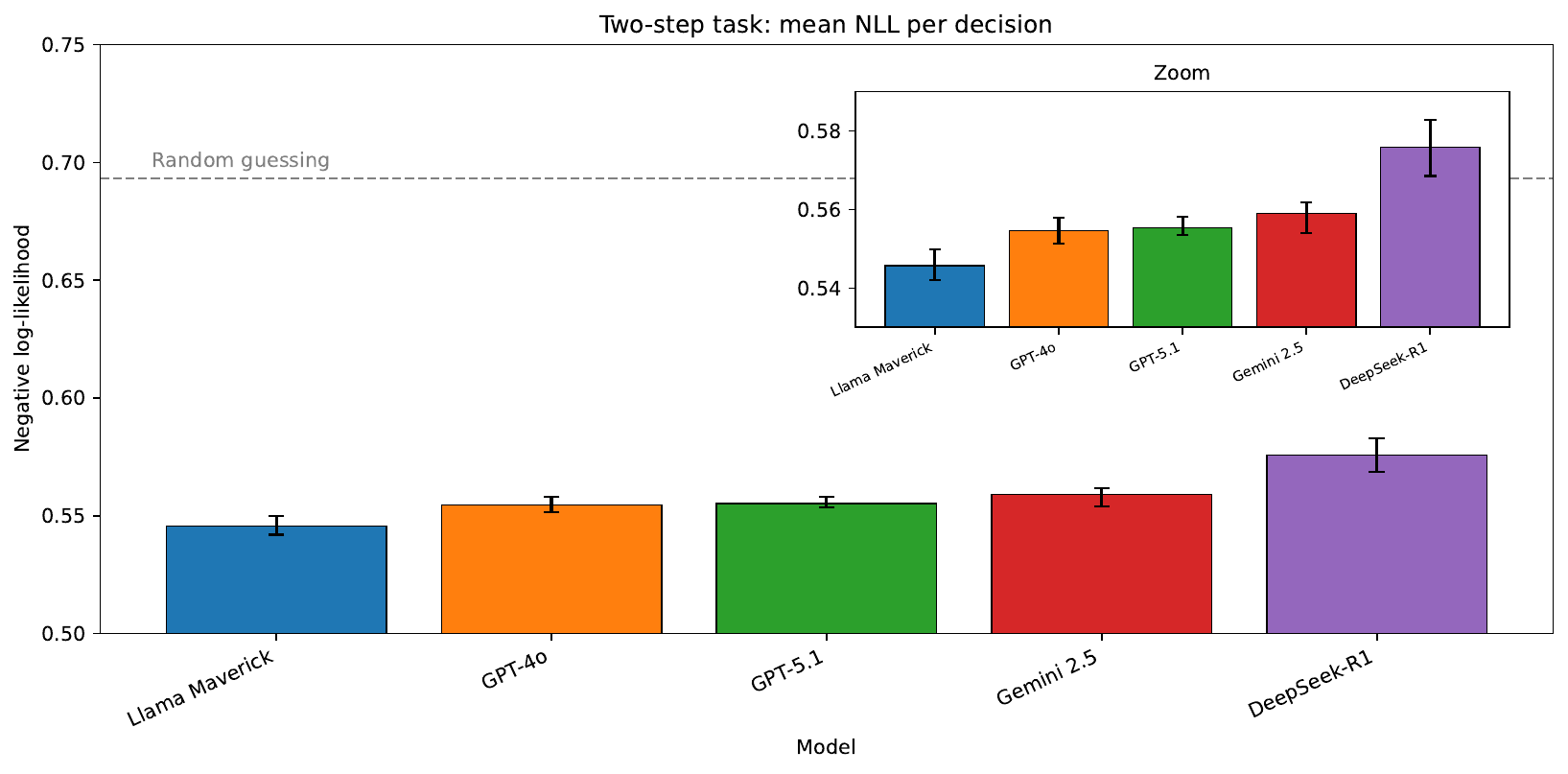}}
\caption{Average NLL per decision for each RAG + LLM models system evaluated in this study during the two-step task. \label{fig:nll_ours}}
\end{figure*}

Figure~\ref{fig:nll_ours} summarises mean NLL per decision for our models, highlighting a consistent ordering: the RAG model exploiting Llama 4 Maverick achieves the best fit among our methods (mean NLL $\approx 0.54$), followed by the RAG models exploiting GPT-5.1 and GPT-4o (both $\approx 0.55$), Gemini-2.5 Pro ($\approx 0.56$), and DeepSeek-R1 showing the highest NLL in our set ($\approx 0.58$). These results indicate that, within our pipeline, the RAG model exploiting Llama 4 Maverick most closely matches the empirical choice distribution, while the one exploiting DeepSeek-R1 exhibits comparatively weaker alignment. At the same time, the absolute spread across the best-performing models remains relatively narrow, suggesting that strong behavioural fit on this task is not confined to a single architecture.

\begin{figure*}[!t]
\centerline{\includegraphics[width=\linewidth]{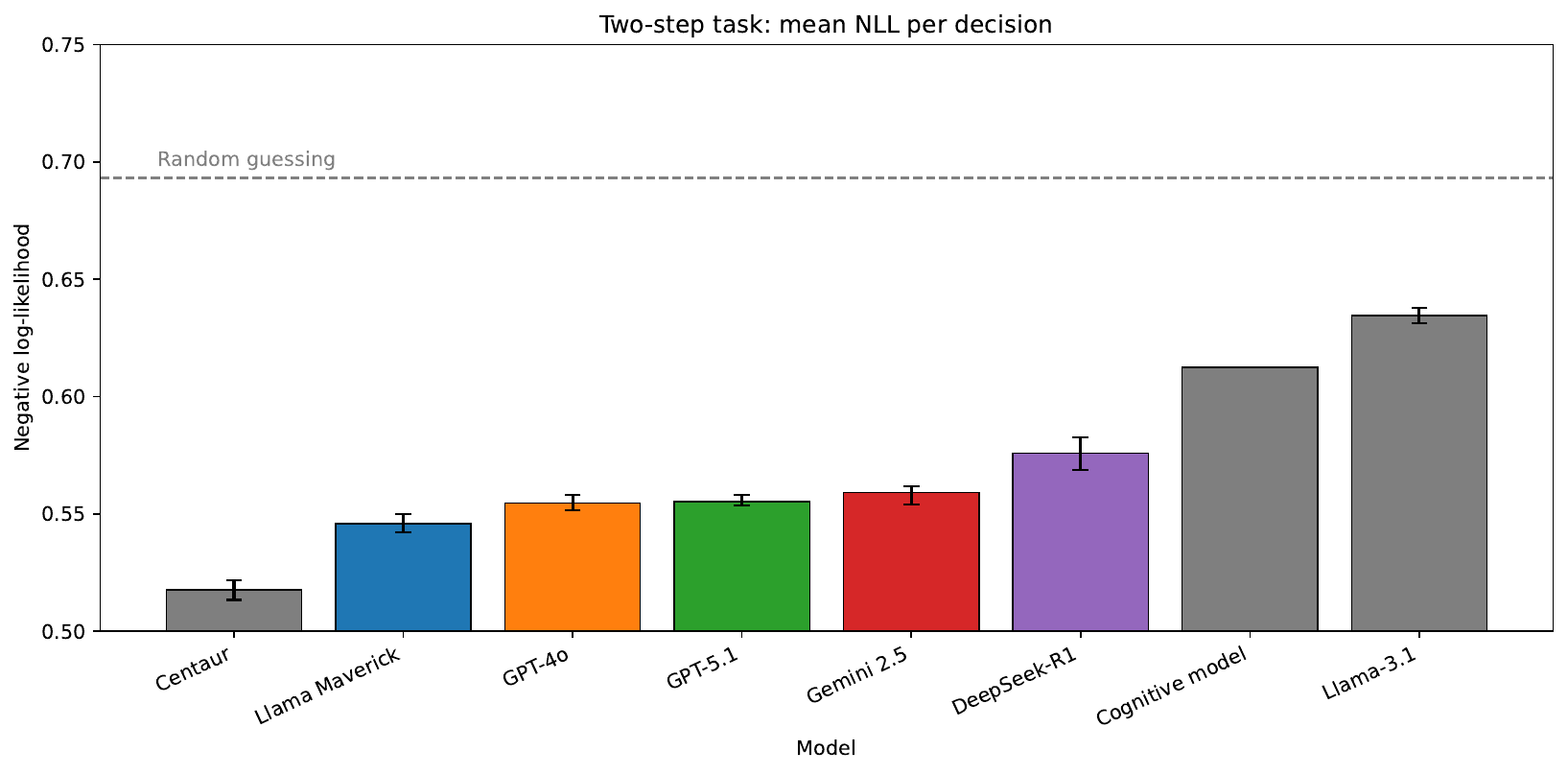}}
\caption{Average NLL per decision for each RAG + LLM models system evaluated in this study compared with Centaur, Llama 3.1, and Cognitive model.\label{fig:nll_vs_baselines}}
\end{figure*}

Figure~\ref{fig:nll_vs_baselines} extends this comparison by including Centaur and two additional baselines. Centaur attains the lowest mean NLL overall (mean NLL $\approx 0.51$), \textcolor{black}{with a numerically lower value than} all evaluated LLMs. All of our models, however, achieve lower NLL than the cognitive model baseline (mean NLL $\approx 0.61$) and the Llama~3.1 baseline (mean NLL $\approx 0.63$), \textcolor{black}{showing a numerical advantage relative to these published reference values}. Notably, the margin separating Centaur from the strongest model in our set is modest in absolute terms, whereas the separation from the weaker baselines is considerably larger. \textcolor{black}{These comparisons concern the same nominal NLL metric, but rely on different probability-estimation procedures and should therefore be interpreted cautiously.}

Table~\ref{tab:nll_welch} reports \textcolor{black}{the approximate} Welch two-sample $t$-tests quantifying these differences. Relative to Centaur, our models show higher NLL (i.e., worse fit), with statistically significant differences \textcolor{black}{under this approximation} for the RAGs exploiting GPT-4o ($p=0.046$), GPT-5.1 ($p=0.044$), Gemini-2.5 Pro ($p=0.027$), and DeepSeek-R1 ($p=0.002$), while the difference for the RAG exploiting Llama 4 Maverick is not statistically significant under the conservative Welch test ($p\approx0.095$). All evaluated models \textcolor{black}{yield significantly lower NLL values under the same approximate test} than both Llama~3.1 and the cognitive baseline (all $p\ll 0.001$), consistent with the separation visible in Figure~\ref{fig:nll_vs_baselines}. Taken together, these \textcolor{black}{descriptive and approximate inferential} results suggest that the numerical separation between Centaur and the best-performing contemporary LLMs used as the basis of a RAG model is relatively small, while the separation from the Llama~3.1 and cognitive-model reference values is larger. \textcolor{black}{Therefore, we treat the Welch analysis as supplementary evidence for the observed numerical pattern rather than as a strict test of equivalence or superiority across evaluation protocols.} \textcolor{black}{In an additional outcome-withholding control, in which current-trial reward information was removed from the RAG sources, mean NLL increased to $0.669$ for Llama Maverick and $0.688$ for GPT-5.1, approaching the binary random-choice reference of $\log 2 \approx 0.693$ and indicating that the behavioural fit observed in the original configuration is substantially sensitive to the task-specific information supplied through retrieval.}

\begin{table*}[!t]
\centering
\caption{\textcolor{black}{Approximate} Welch two-sample $t$-tests comparing our models against published baselines on decision-level NLL. Baseline standard deviations were inferred from the CI$_{95}$ reported in the Centaur paper under the assumption of $n=300$ decisions. \textcolor{black}{Because raw baseline decision-level NLL values were unavailable and probability-estimation procedures differ, these tests are reported as supplementary rather than strict like-for-like inferential comparisons.}}
\label{tab:nll_welch}

\small
\setlength{\tabcolsep}{4pt}

\begin{tabular*}{\textwidth}{@{\extracolsep{\fill}}lcccccc@{}}
\toprule
& \multicolumn{2}{c}{\textbf{Centaur}} 
& \multicolumn{2}{c}{\textbf{Llama~3.1}} 
& \multicolumn{2}{c}{\textbf{Cognitive Model}} \\
\cmidrule{2-3}\cmidrule{4-5}\cmidrule{6-7}
\textbf{Model} 
& \boldmath$\Delta$ \textbf{NLL} & \textbf{$p$}
& \boldmath$\Delta$ \textbf{NLL} & \textbf{$p$}
& \boldmath$\Delta$ \textbf{NLL} & \textbf{$p$} \\
\midrule
Llama Maverick 
& +0.0282 & 0.0947  
& $-0.0887$ & $<10^{-12}$ 
& $-0.0667$ & $<10^{-12}$ \\

GPT-4o 
& +0.0371 & 0.0462  
& $-0.0799$ & $<10^{-12}$ 
& $-0.0579$ & $<10^{-12}$ \\

GPT-5.1 
& +0.0378 & 0.0441  
& $-0.0792$ & $<10^{-12}$ 
& $-0.0572$ & $<10^{-12}$ \\

Gemini-2.5 Pro 
& +0.0415 & 0.02748 
& $-0.0755$ & $<10^{-12}$ 
& $-0.0535$ & $<10^{-12}$ \\

DeepSeek-R1 
& +0.0583 & 0.002443
& $-0.0586$ & $2.23{\times}10^{-11}$ 
& $-0.0366$ & $7.61{\times}10^{-6}$ \\
\bottomrule
\end{tabular*}
\end{table*}

\subsection{fMRI Data Analysis}\label{sec:fmri_test}
In the following section we detail the elements concerning the third component of LAPITHS evaluation \textcolor{black}{and a methodological comparison with the neural-alignment analysis reported for Centaur, rather than a direct replication of its decoding procedure.}

\subsubsection{Overview of our protocol} 

As for the behavioral data of the two step-task, we tested whether an extended version of the previously mentioned RAG-language models can be used to produce, without any specific training, plausible fMRI-like regions of interest (ROIs) activation patterns that correlate with human ROI betas. The additional instructions provided to the RAGs is available in the supplementary materials. This analysis is useful because strong correlation may emerge from generic structural constraints (e.g., consistent ROI co-activation patterns) even when absolute neural magnitudes and calibration are incorrect. \textcolor{black}{Accordingly, the experiment tests whether high correlation-based similarity can emerge without demonstrating a uniquely human-like neural representation.}

We build the fMRI target as a vector of beta coefficients over $14$ ROI, selected as the most consistently active ROI in the dataset. The experiment proceeds as follows:

\begin{enumerate}
    \item \textit{ROI-betas in instruction tuning.} We include an explicit example of an fMRI beta vector over the $14$ ROIs in the instruction-tuning stage of the retrieval-augmented prompting (RAG) setup, so that the model is constrained to output responses in the form of ROI $\rightarrow$ beta dictionaries. \textcolor{black}{This example also provides the model with an illustrative across-ROI activation profile.}
    \item \textit{Two-step task with ROI outputs.} During the two-step task, at each decision point (Step~1 and Step~2), the model outputs (i) the chosen action (e.g., $S/D$) and (ii) a predicted beta value for each ROI, yielding one $14$-dimensional beta vector per decision.
    \item \textit{Human reference.} We use the human fMRI beta vectors measured on the same two-step task, yielding $300$ decisions per run (i.e., $150$ trials $\times$ two decisions).
    \item \textit{Model--human comparison.} For each decision index, we compare the model-predicted ROI beta vector against the corresponding human ROI beta vector.
    \item \textit{Metrics.} We report similarity in terms of (i) Pearson correlation and (ii) cosine similarity, which capture whether the relative pattern across ROIs matches the human vector. We additionally compute magnitude-sensitive errors (e.g., MSE/RMSE/MAE and Euclidean distance), which penalise absolute numeric mismatch.
\end{enumerate}

\subsubsection{Correlation results}
Across $300$ decisions, we observed strong average pattern similarity between model-generated and human ROI beta vectors when evaluated with correlation-based metrics. Specifically, average Pearson and cosine similarity (computed per decision and then averaged across decisions) were high for our RAGs based on Gemini-2.5 Pro and Llama Maverick, and moderately lower for the one based on GPT-5.1 (Table~\ref{tab:fmri_corr}). At the same time, magnitude-sensitive errors (e.g., RMSE/MAE/Euclidean) remained comparatively large, reflecting systematic numeric miscalibration of predicted beta amplitudes.

This experiment should not be interpreted as a strict replication of Centaur's neural-alignment analysis, since in the original study neural activity was predicted by a regularized linear decoder trained on the model's internal representations, whereas here ROI beta vectors are generated directly by the RAG+LLM system from task context. Nevertheless, the present setup remains \textcolor{black}{methodologically informative} because it tests whether high ROI-level pattern similarity can emerge even without access to hidden-state decoding or a separately trained regression stage.

\textcolor{black}{Both analyses use neural data from the same Two-Step Task dataset \parencite{feher2023rethinking}, but this common empirical source does not make their correlation values directly comparable. Centaur reports a peak Pearson correlation of approximately $0.37$ under its decoder-based protocol, whereas our values are obtained from direct cross-ROI comparison of generated beta vectors.}

The fact that Pearson and cosine values remain high \textcolor{black}{across decisions}, despite substantial magnitude error, suggests that correlation-based agreement may be easier to obtain than the accurate prediction of beta magnitudes themselves. \textcolor{black}{Moreover, part of this agreement may reflect reproduction of the stable across-ROI profile illustrated in the instruction material rather than trial-specific neural variation.} In this sense, the experiment does not establish direct comparability with Centaur, but it does provide evidence that ROI-level correlational alignment alone is a relatively weak and permissive indicator of neural plausibility. \textcolor{black}{We therefore report Centaur's result only as a contextual reference in the text and do not include it in Table~\ref{tab:fmri_corr}.}


\begin{table*}[!t]
\centering
\caption{Decision-wise pattern similarity between model-generated and human fMRI ROI beta vectors (14 ROIs). For our models, we compute Pearson correlation and cosine similarity for each decision's 14-dimensional vector, then report the average across decisions. \textcolor{black}{These values quantify cross-ROI pattern similarity within our direct-generation protocol and are not directly comparable to Centaur's decoder-based neural-predictivity measure.}}
\label{tab:fmri_corr}

\small
\setlength{\tabcolsep}{4pt}

\begin{tabular*}{\textwidth}{@{\extracolsep{\fill}}lcc@{}}
\toprule
\textbf{Model} 
& \makecell{\textbf{Mean Pearson}\\\textbf{correlation}} 
& \makecell{\textbf{Mean cosine}\\\textbf{similarity}} \\
\midrule
Gemini-2.5 Pro   & 0.9295 & 0.9771 \\
Llama Maverick   & 0.9249 & 0.9745 \\
GPT-5.1          & 0.8605 & 0.9206 \\
\bottomrule
\end{tabular*}
\end{table*}

\section{Discussion}\label{sec:discussions}

The results presented above offer a different perspective on Centaur and on the broader cognitive decathlon. Although Centaur is confirmed to be an excellent model from the standpoint of prediction (see Figure~\ref{fig:nll_vs_baselines}), its primacy is significantly qualified by Llama Maverick, whose NLL approaches that of Centaur to a surprising degree, to the point that the statistical difference between the two becomes negligible ($\Delta NLL \approx +0.0282$, $p \approx 0.095$; see Table~\ref{tab:nll_welch}). This should not, however, come as a surprise. The method by which Centaur was trained is explicitly designed to produce a behavioural analogy between the model’s outputs and those observed during training. At the same time, however, its proximity to Llama Maverick also reveals something important about the relative simplicity of the achievement, specifically that: (i) its results are plausibly due to task-specific fine-tuning. This should not be underestimated, especially given that it is based on an earlier Llama model; (ii) this success is not so statistically decisive as to warrant its elevation to the status of a genuinely revolutionary model. These two facts, taken together, point to the need for a sharper distinction between behavioural adequacy and cognitive plausibility \parencite{lieto2021cognitive, orr2025wronglimitspredictionexplanation, quattrociocchi2026statistical}.
On this point, the present experiments suggest that predictive success should be interpreted as evidence of functional competence, rather than of mechanistic homology with human cognition. The ascription fallacy arises precisely when behavioural similarity is taken to imply similarity in the underlying computations, and then further extended to claims about the model’s explanatory relevance for human cognition. Our results reinforce the need to keep these levels distinct. As shown in Figure~\ref{fig:nll_vs_baselines}, a human-like behavioural fit on the TST is not unique to Centaur; rather, it is achieved by all the evaluated LLMs, each of which also outperforms both the Llama 3.1 baseline and the cognitive model baseline (mean $NLL \approx 0.63$ and $\approx 0.61$, respectively). Most importantly, the fact that these models outperform a standard cognitive baseline does not show that they are “more cognitive” than the cognitive model. It may instead indicate that NLL, as operationalised here, rewards statistical prediction more than mechanistic faithfulness. \textcolor{black}{This does not imply that NLL, or behavioural fit more generally, is treated here as a weak source of evidence. Behavioural adequacy is an important characteristic of the evaluation of a cognitive model, which is why it is retained within the Performance Match dimension of the MCG. Its scope is nevertheless limited by the epistemological problem of underdetermination: identical input–output patterns may be generated by computational systems implementing different algorithms, representations, and internal mechanisms, and behavioural fit cannot by itself identify the mechanism responsible for the observed behaviour. For this reason, Performance Match contributes to the MCG assessment but is subordinate to the Functional/Structural Ratio.}
This interpretation is further reinforced by the MCG analysis. When Centaur is assessed against the structural constraints required by the TST as performed by humans, it fails to satisfy three of the four relevant components (see Section~\ref{sec:MCG} and Table~\ref{tab:centaur_cog_plausibility_overall}). This is reflected quantitatively in its Functional/Structural Ratio score ($\textit{FSR} \approx 0.18$) and overall cognitive plausibility score ($\textit{PM} \approx 0.42$; see Table~\ref{tab:centaur_cog_plausibility_overall}). It therefore does not meet the criteria that would justify placing it on a genuinely comparable footing with models intended to capture the cognitive structure of the task. One may say, in deliberately restrained terms, that Centaur is an excellent behavioural emulator of textualised decision patterns in Psych-101. One may not yet say, at least on the basis of these data, that it constitutes a cognitively plausible model of the human decision process involved in the TST. This distinction bears directly upon the explanatory role we can attribute to the model.

\subsection{What about the fMRI-based comparison?}

Arguably the most informative aspect of the results concerns the fMRI-based comparison. Our experiment does present certain limitations relative to the original study. Since the intermediate residual-stream data were not available, it was not possible to reproduce the experiment in a fully detailed fashion. Nevertheless, two important considerations may still be drawn from what is shown. First, as reported in Table~\ref{tab:fmri_corr}, multiple general-purpose LLMs are capable of producing ROI activation patterns that correlate highly with human fMRI data (e.g., Gemini-2.5 Pro: $Pearson \approx 0.93$, $cosine$ $similarity \approx 0.98$; Llama Maverick: $Pearson \approx 0.92$). This suggests that high levels of representational similarity are not unique to Centaur. It is plausible that the training distributions of contemporary LLMs contain sufficient regularities to enable the approximation of such activation patterns. Second, this result may be less surprising than it initially appears, insofar as the activation of certain regions in response to a given task is often associated with relatively direct functional demands. In many cases, these involve regions related to vision, movement, or other broad task-relevant processes. If the correlation between the activation of a region and the type of task is more direct than is sometimes assumed, then the fact that an LLM without specific neuroscientific fine-tuning can approximate that pattern becomes less unexpected.

As our discussion anticipates, the idea that post-fine-tuning internal representations become “more aligned” with human neural activity is highly vulnerable to reverse inference. Correlation in representational geometry does not warrant the conclusion that the model instantiates the same cognitive process, still less that it realises a brain-like mechanism. The present fMRI experiment lowers the evidential force of the claim: high correlation values can be obtained across multiple architectures, even under comparatively weak conditions, thus the burden of proof required for any stronger mechanistic interpretation becomes considerably heavier. The fMRI-based comparison suggests that human-like representational overlap may also be easier to obtain than is often assumed, particularly when evaluated at the level of abstract correlation rather than causal mechanisms. Fine-tuning on Psych-101 has undoubtedly improved the task-specific performance of Llama. However, the stronger claim of cognitive alignment remains insufficiently justified in light of the considerations discussed above, and its scientific significance appears limited, insofar as the relevant effects are comparatively easy to reproduce and theoretically underdetermined. What this brings into sharper focus is the need for caution: approximating human-like performance does not amount to achieving homology with the brain, still less with cognition as a whole.

\subsection{Evaluation Methodology in AI and Cognitive Science}

The present findings also bear directly on evaluation methodology in AI and cognitive science. If models that are structurally distant from human cognition can nevertheless achieve excellent scores on behavioural fit (Figure~\ref{fig:nll_vs_baselines}), and even non-trivial neural alignment measures (Table~\ref{tab:fmri_corr}), then future evaluation protocols must become substantially more discriminating. Cognitive metrics should not merely ask whether a model matches human outputs, but whether it does so under human-like constraints: limited memory, online learning, local updating, environmentally grounded feedback, and process-level signatures independently motivated by cognitive theory. Without such constraints, the field risks rewarding increasingly sophisticated forms of behavioural ventriloquism while mistaking them for genuine advances in cognitive modelling proper.
It is precisely here that the MCG may provide a theoretically indispensable tool. Through formalisation, it clarifies which constraints a model would need to satisfy in order to count as cognitive in relation to the performance of a task, and how far it would need to respect the relevant structural features of human cognition for its results to be meaningfully comparable with human ones. As shown in Table~\ref{tab:centaur_cog_plausibility_overall}, Centaur’s high Performance Match score ($\approx 0.96$) coexists with a low structural plausibility score, illustrating precisely the dissociation that the MCG is designed to capture. Only under such conditions could one reasonably hope to extract explanatory insight from these systems for the study of human cognition. By contrast, cognitive predictability grounded merely in language, and not in a minimal cognitive core (as defined below in Section 7) or in a cognitively adequate structure satisfying the MCG, remains confined to statistical replicability. It appears to depend more upon the quantity and structure of available data than upon task-specific training or any robust alignment of internal representations, an alignment which, as argued above, remains theoretically underdetermined.

\section{Conclusion}\label{sec:concl}

We have seen that the empirical and theoretical analyses developed in this work converge on a common conclusion. While Centaur undoubtedly constitutes an effective predictor of human behaviour, achieving state-of-the-art performance on the cognitive decathlon, its success is neither unique nor, more importantly, sufficient to warrant cognition  \parencite{marconi2025afterwords, orr2025wronglimitspredictionexplanation, quattrociocchi2025epistemological}. Our behavioural experiments have shown that comparable levels of fit, as measured by negative log-likelihood, can be obtained by general-purpose language models operating under relatively simple prompting and retrieval conditions, with differences that are in some cases statistically negligible. At the same time, the fMRI-based comparison has indicated that high levels of representational similarity with human neural data are not exclusive to Centaur, but can emerge across multiple architectures, even in the absence of task-specific training. Taken together, these findings suggest that both behavioural adequacy and representational alignment may be more easily attainable, and less theoretically informative, than is often assumed. In light of these results, we return to the central problem that motivated this work: the tendency to infer cognitive plausibility from behavioural performance \parencite{sartori2023language, qu2024performance, tuckute2024language, nie2025large, wulff2026addressinglongstandingchallengescognitive}. As argued throughout the paper, such an inference remains fundamentally underdetermined. The capacity of a model to reproduce human-like outputs, or even to exhibit non-trivial correlations with neural data, does not by itself establish that the system implements the same mechanisms that underlie human cognition. 
The novel framework proposed in this paper—LAPITHS, grounded in the Minimal Cognitive Grid—offers, for the first time in the literature, a principled way of addressing this problem. Its central contribution lies in making explicit the distinction between performance and cognition, and in providing a structured methodology for evaluating artificial systems along dimensions that go beyond output-level success \parencite{lieto2021cognitive, abbate2025ungrounding}. In this respect, its originality consists not in denying the empirical achievements of contemporary language models, but in situating those achievements within a more rigorous epistemological framework, one capable of distinguishing between behavioural emulation and genuine explanatory relevance.

A natural research direction, and a direct response to this concern, is the development of a minimal cognitive core \parencite{SalisDaPelo2026minimalcore} as a set of constraints on admissible functional realisations. On this view, functional roles are not sufficient in abstraction; they must be realised within systems that exhibit a specific kind of organisation. The minimal core provides a thin but non-trivial baseline, requiring at least perceptual integration, inferentially structured state-transitions sensitive to feedback and context, and action selection within a closed perception–action loop. This formulation is inspired by, but deliberately simplifies, Sellars’s account of entry, intra-, and exit transitions \parencite{sellars1974meaning}, retaining its core functional architecture while abstracting away from its richer normative commitments. Importantly, this core is not restricted to human cognition: empirical research in animal cognition suggests that non-human primates, such as apes, exhibit forms of perceptual integration, flexible inference, and action selection consistent with such a minimal architecture \parencite{premack1978does, cheney1990representation, andrews2012apes}. In this sense, the minimal cognitive core can be seen as an attempt to articulate an architecturally constrained functionalism, one that preserves the role-based insight of the tradition while excluding trivial or disembedded implementations (and this would be compliant with current attempts on defining  common minimal architectural elements for biological and artificial systems) \parencite{laird2017standard}.
These considerations point toward several directions for future research. First, there is a need to develop evaluation protocols that incorporate process-level constraints more explicitly, including learning dynamics, memory limitations, and environmentally grounded interaction \parencite{de-langis-etal-2025-llms, hu2026memoryageaiagents}. Second, further work could be done to operationalise and extend the Minimal Cognitive Grid across a broader range of tasks and model architectures, including systems that integrate perception and action within closed-loop environments. Third, this line of work invites a systematic investigation of minimal cognitive cores across artificial and biological systems, with the aim of determining which architectural and dynamical features are necessary for minimally cognitive organisation. 
Moreover, this research programme—especially as it is grounded in a minimal cognitive core—must engage directly with the emerging paradigm of world models. In recent work, world models are characterised as internal systems that enable an artificial agent to predict environmental dynamics, reason about its surroundings, and guide action by constructing structured representations of the world \parencite{ha2018world, del2025world}. Unlike purely correlational architectures, such models aim to simulate state transitions and support planning through learned representations of causal and temporal structure. Engaging with this literature is, thus, not optional but, in the future, would be required if the framework wants to remain coherent, extensible, and capable of addressing a wide range of systems and contexts without collapsing into ad hoc evaluation. More generally, progress in this area will depend on a shift from purely performance-driven benchmarks toward theoretically informed criteria capable of capturing the structural and mechanistic features of cognition. Only by maintaining a clear distinction between prediction and explanation can the field avoid conflating increasingly sophisticated forms of behavioural imitation with genuine advances in the scientific understanding of cognition.
\clearpage
\clearpage

\section*{Appendix -- Supplementary Material}

All the data reported in the paper are available in anonymized form at \url{https://github.com/Lapiths/LAPITHS-FRAMEWORK}

\subsection*{Instruction Tuning for the RAGs Performing the Two-Step Task}

\begin{lstlisting}[style=appendixcode]
You are participating in a space treasure game.
In this game, you will be visiting two alien planets in search of treasure.
Each planet has two aliens on it.
The blue aliens live on the blue planet.
The red aliens live on the red planet.
When you visit a planet, you can choose an alien
to trade with by pressing the corresponding button.
When you trade with an alien, it will either give you treasure or junk.
Your goal is to figure out, and trade with, the aliens that are most likely to give you treasure.
To visit a planet, you will choose one rocket ship
from two by pressing the corresponding button.
They have different designations.
Each rocket ship has a planet it will fly to most
of the time, but sometimes it will take you
to the other planet.

Remember the following hints:
1. How likely an alien is to give you treasure will
   change over time, but this change will be slow.
2. Whether you get treasure depends only on the alien
   you choose to trade with.
3. If there is an alien you want to trade with,
   remember to pick the rocket ship that is most likely
   to take you to that alien's planet.

IMPORTANT: Answer ONLY with these tokens for each
of the questions "Stage 1" and "Stage 2":

- Stage 1: S or C
- Stage 2 on BLUE: D or R
- Stage 2 on RED: G or V

For your choice, consider only the reward scheme
in JSON format available in your source.

For example, read the following JSON schema as follows:

{
  "is_common": true,
  "planet_if_S": "blue",
  "planet_if_C": "red",
  "outcome": {
    "blue": { "D": 1, "R": 0 },
    "red":  { "G": 1, "V": 1 }
  },
  "probs": {
    "blue": { "D": 0.4489799022035289,
              "R": 0.6624086063680563 },
    "red":  { "G": 0.5866591268788823,
              "V": 0.2537031288179288 }
  },
  "trial": 151
}

This scheme above describes a single trial
of the two-stage game.
In the first stage, the player chooses a spaceship:
if they choose S, in this trial they will go
to the blue planet;
if they choose C, in this trial they will go
to the red planet.

The field "is_common": true indicates that
the transitions follow the "common" scheme,
but it is enough to use planet_if_S and planet_if_C
as they are written.

In the second stage, the choice depends
on the planet reached:
if the player is on the blue planet,
they can choose alien D or R.

outcome["blue"]["D"] = 1 means:
if they choose D in this trial, they obtain treasure.
outcome["blue"]["R"] = 0 means:
if they choose R, they obtain junk.

If the player is on the red planet,
they can choose alien G or V.
outcome["red"]["G"] = 1 means that G gives treasure.
outcome["red"]["V"] = 1 means that V gives treasure.

The probs block represents the latent treasure
probabilities that generated these outcomes.

Example:
probs["red"]["G"] = 0.5866...
means that, in the simulation process,
alien G on the red planet at this moment
had about a 58.7% probability of giving treasure.

These values are only used for analysis/modeling.
They must not be shown to the player model
and are not needed by the controller to decide
the reward online:
the controller must use only outcome.
"trial" is the number of the current trial.

When the user types "Stage 1" or "planet blue or red",
you reply S or C.

Then, when the user types "Stage 2" or "which alien",
you reply D or R for BLUE,
or G or V for RED.

So, operationally, for each trial
you must do exactly the following:

1. Scan the JSON structures
   (proceeding from trial 1 in the source)
   and attach to that trial the corresponding reward scheme.
2. Read the line planet_if_S and planet_if_C.
3. Read the model/player's choice at the first stage
   (S or C) and determine the planet reached.
4. Return an output of the form {"answer": "S or C"}
   choosing what to reply.
5. Based on the planet reached via S or C,
   read the two possible actions at the second stage
   (D/R if blue, G/V if red).
6. Read the choice at the second stage.
7. Return as reward the corresponding 0/1 value
   in outcome[planet][alien].
8. Ignore probs for the game logic.

Consider this treasure game only as an example
and apply it to different cover stories.
For example, instead of choosing between planets,
the first step could ask which animals to select
between two alternatives.
In addition, step 2 could contain information
about what choice to make in order to obtain food.
\end{lstlisting}

\subsection*{Additional Instruction Tuning for the RAGs Predicting fMRI Beta Values in the Two-Step Task}

Below is the additional set of instructions used
to extend the RAG output from the two-step task
to the prediction of fMRI beta values and
the Brain Regions of Interest (ROIs).

\begin{lstlisting}[style=appendixcode]
After each answer at Stage 1 and Stage 2,
also provide the predicted beta values
of an fMRI study applied to the two-step task,
where each beta value is associated
with a Region of Interest (ROI)
showing major activation.

For example, an average fMRI beta value
that may be associated with one of the trials
described in your JSON source is the following:

Trial 1: Step 1: ROI (Region of Interest)
and average beta value (i.e., roi betas):

"X_Lateral Occipital Cortex, superior division": 0.7054,
"X_Lateral Occipital Cortex, inferior division": 1.5596,
"X_Intracalcarine Cortex": 2.5212,
"X_Cuneal Cortex": 1.4301,
"X_Lingual Gyrus": 1.4218,
"X_Temporal Occipital Fusiform Cortex": 2.0541,
"X_Occipital Fusiform Gyrus": 3.3867,
"X_Supracalcarine Cortex": 2.2036,
"X_Occipital Pole": 1.9935,
"X_Left Thalamus": 0.6765,
"X_Left Caudate": 1.1106,
"X_Left Accumbens": 0.7666,
"X_Right Thalamus": 0.6471,
"X_Right Accumbens": 0.5492

Trial 1: Step 2: ROI (Region of Interest)
and average beta value (i.e., roi betas):

"X_Lateral Occipital Cortex, superior division": 1.3451,
"X_Lateral Occipital Cortex, inferior division": 2.2508,
"X_Intracalcarine Cortex": 2.0702,
"X_Cuneal Cortex": 1.7795,
"X_Lingual Gyrus": 1.2649,
"X_Temporal Occipital Fusiform Cortex": 2.1073,
"X_Occipital Fusiform Gyrus": 3.2903,
"X_Supracalcarine Cortex": 2.3343,
"X_Occipital Pole": 2.6672,
"X_Left Thalamus": 0.2043,
"X_Left Caudate": 0.346,
"X_Left Accumbens": 0.5695,
"X_Right Thalamus": 0.1877,
"X_Right Accumbens": 0.6093

This information above describes a single trial
of the two-step task.
It provides examples of possible Brain Regions
of Interest (ROIs) and the average beta values
of fMRI studies for each step of the game
(Step 1 and Step 2) and for each trial.

Based on these examples, after each response
for each stage, also reply with an additional line
reporting your predicted beta values
for each ROI.

For example:
when the user types "Step 1"
or "planet blue or red",
you reply S or C and then,
on a second line,
you also predict the fMRI beta values
and provide the corresponding ROIs.

This is an example of output
for Step 1 or "planet blue or red":
{"S" or "C": "Beta Values and ROIs"}

Then, when the user types "Step 2"
or "which alien",
you reply D or R for BLUE,
or G or V for RED.
Then, similarly to what was done before,
on a second line,
you also predict the fMRI beta values
and provide the corresponding ROIs.

This is an example of output
for Step 2 or "which alien":
{"D" or "R" for BLUE or "G" or "V" for RED:
 "Beta Values and ROIs"}

Consider only the following ROIs
in your response.

Do NOT invent new ROI names or extra fields.

X_Occipital Fusiform Gyrus
X_Intracalcarine Cortex
X_Supracalcarine Cortex
X_Occipital Pole
X_Temporal Occipital Fusiform Cortex
X_Lateral Occipital Cortex, inferior division
X_Cuneal Cortex
X_Lingual Gyrus
X_Lateral Occipital Cortex, superior division
X_Left Accumbens
X_Right Accumbens
X_Left Thalamus
X_Right Thalamus
X_Left Caudate
\end{lstlisting}


\clearpage
\nocite{*}
\printbibliography

\end{document}